\documentclass{article}

\newif\ificml
\icmlfalse

\pdfoutput=1
\usepackage[utf8]{inputenc}
\usepackage[T1]{fontenc}
\usepackage{microtype}
\usepackage{graphicx}
\usepackage{subfigure}
\usepackage{booktabs}
\usepackage{hyperref}

\usepackage{url}
\usepackage{nicefrac}
\usepackage{xr-hyper}
\usepackage{listings}
\usepackage{mathtools}
\usepackage{amsfonts}
\usepackage{amssymb}
\usepackage{amsmath}
\usepackage{amsthm}
\usepackage{docmute}
\usepackage[acronym, nomain]{glossaries}

\ificml
\usepackage{icml/icml2019}  % \usepackage[accepted]{icml/icml2019}
\else
\usepackage{natbib}
\usepackage{algorithm,algorithmic}
\fi

\DeclareMathOperator*{\expectation}{\mathop{\mathbb{E}}}
\DeclarePairedDelimiterX{\KL}[2]{\mathrm{KL}[}{]}{#1\;\delimsize\|\;#2}

\DeclarePairedDelimiterX\braket[2]{\langle}{\rangle}{#1 \delimsize\vert #2}

\newcommand{\prior}{{\mathbb{P}}}
\newcommand{\posterior}{{\mathbb{Q}}}
\newcommand{\train}{{\mathcal{S}}}
\newcommand{\true}{{\mathcal{D}}}

\newcommand{\expect}[2]{\expectation_{#1}\left[#2\right]}
\newcommand{\lossfun}[1]{\mathcal{L}_{#1}}
\newcommand{\loss}[2]{\lossfun{#1}\left({#2}\right)}
\renewcommand{\vec}[1]{{\boldsymbol{#1}}}
\newcommand{\param}{{\vec{\theta}}}

\newcommand{\spectral}[1]{\|{#1}\|_2}
\newcommand{\frobenius}[1]{\|{#1}\|_\mathrm{F}}

\newcommand{\sample}{{z}}

\newcommand{\Sec}[1]{{Sec.\,\ref{#1}}}
\newcommand{\hyp}{{f}}

\newcommand{\iid}{{\textit{i.i.d.}}}

\newcommand{\const}{\mathrm{const.}}
\newcommand{\zoloss}{$0\mathchar`-1$ loss}
\newcommand{\idmat}{I}
\ificml
\newcommand{\Appendix}[1]{{Appendix\,\ref{#1} in the supplementary material}}
\else
\newcommand{\Appendix}[1]{{appendix\,\ref{#1}}}
\fi

\ificml
\usepackage{xr}
\fi

\ificml
\icmltitlerunning{Normalized Flat Minima}
\fi

\begin{document}

\ificml
  \twocolumn[
    \icmltitle{Normalized Flat Minima: Exploring Scale Invariant Definition\\of Flat Minima for Neural Networks using PAC-Bayesian Analysis}
    
    \begin{icmlauthorlist}
    \icmlauthor{Yusuke Tsuzuku}{utokyo1,riken}
    \icmlauthor{Issei Sato}{utokyo2,riken}
    \icmlauthor{Masashi Sugiyama}{riken,utokyo2}
    \end{icmlauthorlist}
    
    \icmlaffiliation{utokyo1}{Department of Computer Science, University of Tokyo, Tokyo, Japan}
    \icmlaffiliation{utokyo2}{Department of Complexity Science and Engineering, University of Tokyo, Tokyo, Japan}
    \icmlaffiliation{riken}{RIKEN, Tokyo, Japan}
    
    \icmlcorrespondingauthor{Yusuke Tsuzuku}{tsuzuku@ms.k.u-tokyo.ac.jp}
    
    \icmlkeywords{Generalization,PAC-Bayes,Flat Minima,Deep Learning}
    
    \vskip 0.3in
  ]
  \printAffiliationsAndNotice{}
\else
  \title{Normalized Flat Minima:\\Exploring Scale-Invariant Definition\\of Flat Minima for Neural Networks\\using PAC-Bayesian Analysis}
  \author{
      Yusuke Tsuzuku\thanks{The University of Tokyo, RIKEN, \texttt{tsuzuku@ms.k.u-tokyo.ac.jp}}\\
      \and
      Issei Sato\thanks{The University of Tokyo, RIKEN, \texttt{sato@k.u-tokyo.ac.jp}}\\
      \and
      Masashi Sugiyama\thanks{RIKEN, The University of Tokyo, \texttt{sugi@k.u-tokyo.ac.jp}}
  }
  \date{}
  \maketitle
\fi

  % ======
  % TODO(tsuzuku)
  % * shorter sentence
  % * make row/column wise to theorem
  % * figure of row-column wise prior
  % * create a section concerning transformations by Dinh et al.
  % * beta dist.
  % ======

  \begin{abstract}
    The notion of flat minima has played a key role in the generalization studies of deep learning models.
    However, existing definitions of the flatness are known to be sensitive to the rescaling of parameters.
    The issue suggests that the previous definitions of the flatness might not be a good measure of generalization, because generalization is invariant to such rescalings.
    In this paper, from the PAC-Bayesian perspective, we scrutinize the discussion concerning the flat minima and
    introduce the notion of normalized flat minima, which is free from the known scale dependence issues.
    Additionally, we highlight the scale dependence of existing matrix-norm based generalization error bounds similar to the existing flat minima definitions.
    Our modified notion of the flatness does not suffer from the insufficiency, either, suggesting it might provide better hierarchy in the hypothesis class.
  \end{abstract}

  \section{Introduction}
  \label{sec:introduction}

    Theoretical understanding of the high generalization ability of deep learning models is a crucial research objective for principled improvement of the performance.
    Insights and formulations are also useful to compare between models and interpret them.
    Some prior work has explained the generalization by the fact that trained networks can be compressed well\,\citep{CompressionBound,MDLofDeep},
    while others have tried to explain it by the scale of the network and prediction margins\,\citep{SpectralMargin,PACBayesianSpectralMargin}.
    From the minimum description length principle\,\citep{MDL}, models representable with a smaller number of bits are expected to generalize better.
    Bits-back arguments\,\citep{KeepNNSimple,BitsBack} have shown that when models are stable against noise on parameters, we can describe models with fewer bits.
    These arguments have motivated the research on ``flat minima.''
    Empirical work has supported the usefulness to measure the flatness on local minima\,\citep{LargeBatchTraining,HessianAnalysis}.
    Other work has proposed training methods to search for flatter minima\,\citep{FlatMinima,EntropySGD,TrainLonger}.
    As measures of the flatness, prior work proposed the volume of the region in which a network keeps roughly the same loss\,\citep{FlatMinima}, the maximum loss around the minima\,\citep{LargeBatchTraining}, and the  spectral norm of the Hessian\,\citep{HessianAnalysis}.

    Despite the empirical connections of ``flatness'' to generalization, existing definitions of it suffer from scale dependence issues.
    \citet{SharpMinimaGeneralize} showed that we can arbitrarily change the flatness of the loss landscape for some networks without changing the functions represented by the networks.
    Such scale dependence appears in networks with ReLU activation functions or normalization layers such as batch-normalization\,\citep{BatchNorm} and weight-normalization\,\cite{WeightNorm}.
    Since generalization does not depend on rescalings of parameters, the scale dependence issues suggest that the prior definitions of ``flatness'' might not be a good measure of the generalization of neural networks.
    The literature showed that solely looking for the flatness is not sufficient and both the flatness and the scale of parameters should be small for generalization\,\citep{Nonvacuous,ExploreGeneralization}.
    However, how to unify the two quantities has been left as an open problem.
    % TODO(tsuzuku) できればここに一文欲しい

\begin{figure}[t]
      \begin{minipage}{.49\hsize}
        \begin{center}
          \includegraphics[width=\linewidth]{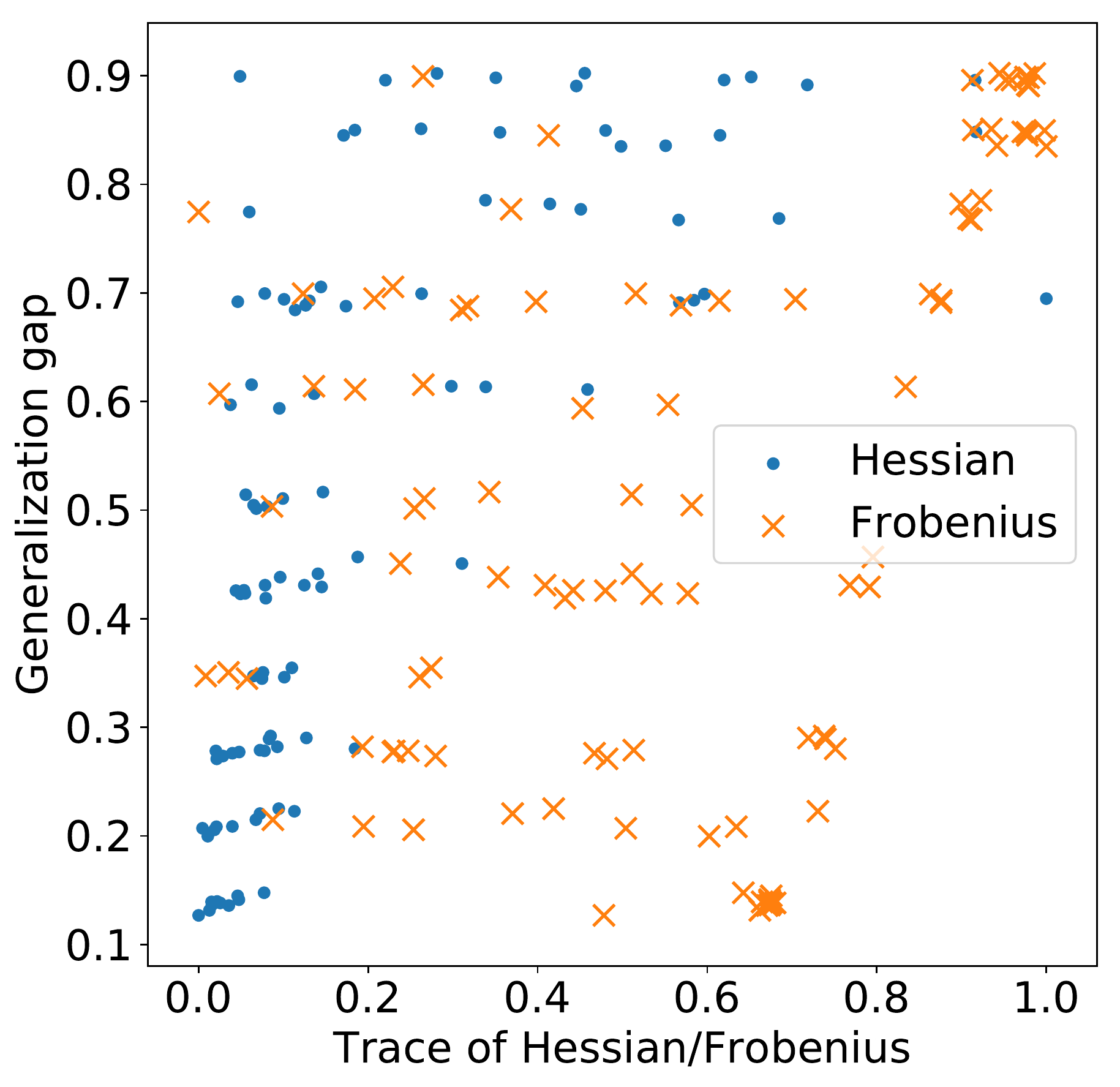}
        \end{center}
      \end{minipage}
      \begin{minipage}{.49\hsize}
        \begin{center}
          \includegraphics[width=\linewidth]{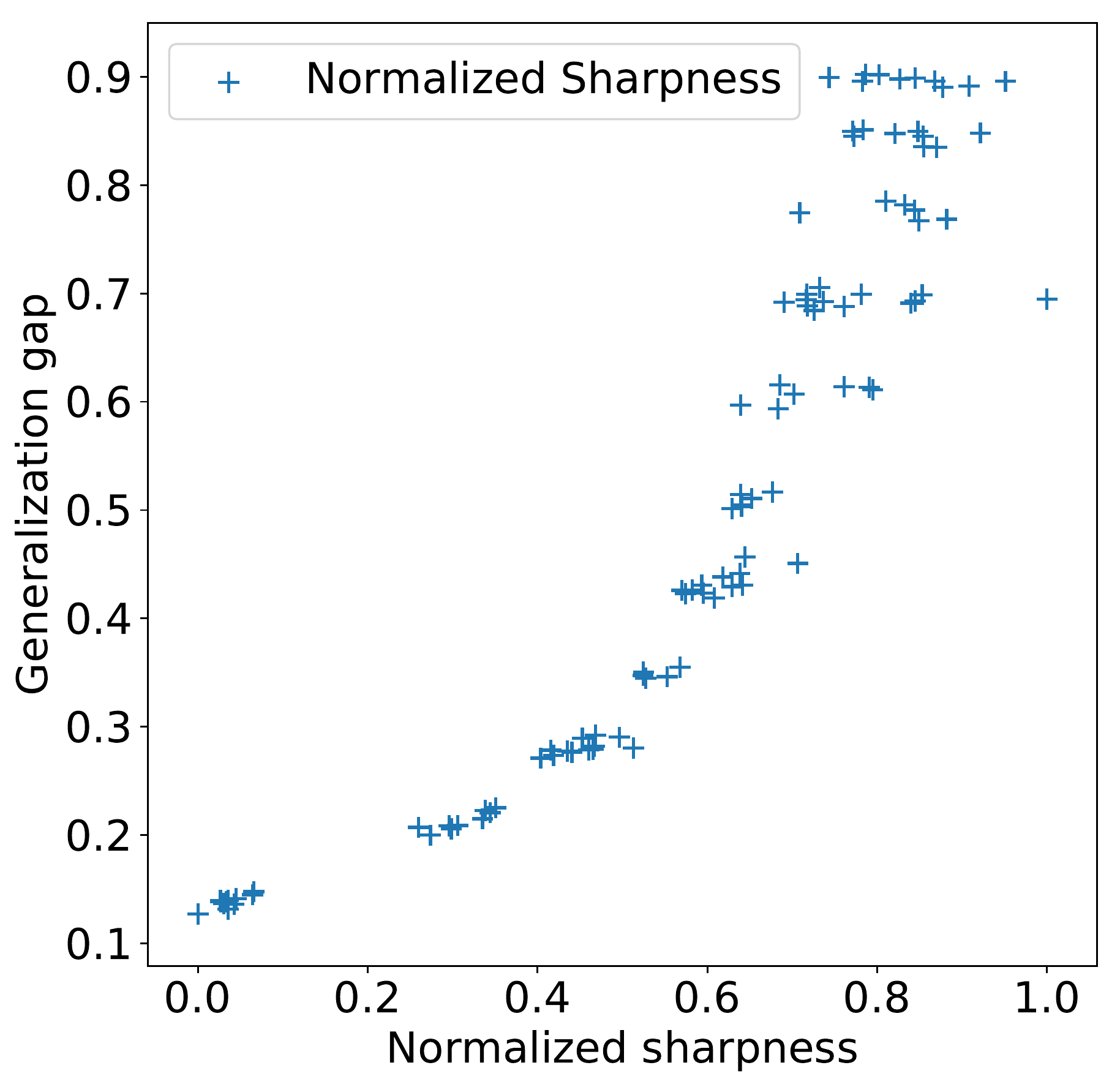}
        \end{center}
      \end{minipage}
      \caption{
        Scatter plot between sharpness measures and generalization gap for Wide ResNets trained on CIFAR10 with different ratio of random labels (\Sec{sec:experiments}).
        The left figure uses the trace of the Hessian of parameters\,\eqref{eq:trace-hessian} and sum of the squared Frobenius norm of weight matrices as the existing sharpness metrics.
        The right figure uses our proposed sharpness metric: normalized sharpness\,\eqref{eq:normalized-flat-minima}.
        All sharpness measures were rescaled to $[0, 1]$ by their maximum and minimum among the trained networks.
        Our modification of the notion of flat minima provides stronger correlation between the sharpness on local minima and the generalization gap.
        The correlation coefficient were $70.8$ and $91.1$ for the trace of the Hessian and the normalized sharpness, respectively.
      }
      \label{fig:intro-fig}
    \end{figure}

    % TODO(tsuzuku) できればmatrixごとのrescalingを組み込んだprior warkに配慮したい
    What causes the problems in the previous definitions of ``flatness?''
    In prior definitions, they implicitly used Gaussian priors with the same variance for all parameters\,(\Sec{sec:flat-minima-pac-bayes}).
    However, as \citet{SharpMinimaGeneralize} pointed out, an assumption that all parameters have the same scale is not good prior knowledge for neural networks.
    In this paper, using a PAC-Bayesian framework\,\citep{SomePACTheorems,PACBayesStochasticModelSel}, we explicitly take scale properties of neural networks into consideration.
    We first incorporate the knowledge that each weight matrix can have different scales\,(\Sec{sec:matrix-normalized-flat-minima}).
    Next, we extend the analysis to row and column wise scaling of parameters\,(\Sec{sec:normalized-flat-minima}).
    To the best of our knowledge, our analysis provides the first rescaling invariant definition of the flatness of local minima.
    Figure\,\ref{fig:intro-fig} shows that our definition of flat minima could distinguish models trained on random labels even when a previous definition fails.

  \section{Related work}
  \label{sec:related-work}

    \citet{Nonvacuous} introduced the PAC-Bayesian framework to study the generalization of deep learning models.
    They also connected PAC-Bayesian arguments to flat minima.
    They pointed out that sharpness is not sufficient and we need to pay attention to the scale of parameters.
    However, they could not remove the scale dependence.
    \citet{ExploreGeneralization} extended \citet{Nonvacuous} and suggested that the sharpness is better to be scaled by the scale of parameters.
    However, they left a way to combine the sharpness and the scale of parameters as an open problem.

    \citet{PACBayesianSpectralMargin} analyzed the generalization of deep learning models using the PAC-Bayesian framework.
    They focused on bounding the worst-case propagation of perturbations.
    Given the existence of adversarial examples\,\citep{Intriguing}, this approach inevitably provides a loose bound.
    Alternatively, we rely on flat-minima arguments, which better capture the effect of parameter perturbations.
    We point out an additional insufficiency of their bound in \Sec{sec:normalized-flat-minima:insufficiency}.
    Our redefined notion of flat minima is free from this issue, suggesting it better captures generalization.

    \citet{IdentifyGen} examined a better choice of the posterior variance and showed that Hessian-based analysis relates to the scale of parameters.
    However, their argument could not overcome the scale dependence.
    Moreover, their analysis is based on a parameter-wise argument, which involves a factor that scales with the number of parameters, making their overall bound essentially equivalent to naive parameter counting.
    In contrast, our analysis completely removes the scale dependence.
    Additionally, our analysis does not have a constant that scales with the number of parameters.

    \citet{VisualizeLoss} demonstrated that normalizing the loss landscape by the scale of filters in convolutional neural networks provides better visualization of the loss landscape.
    While their work empirically showed the effectiveness to normalize the flatness by the scale of parameters, they did not provide theoretical justification why the normalization is essential.
    We provide some theoretical justifications to focus on the normalized loss landscape through the lens of the PAC-Bayesian framework.

    \begin{table}[t]
        \caption{
          Notation table.
        }
        \label{table:notation}
        \small
        \begin{center}
            \begin{tabular}{ l }
                \toprule
                $\prior$: prior distribution of hypothesis (parameters)\\
                $\posterior$: posterior distribution of hypothesis (parameters)\\
                $\true$: underlying (true) data distribution\\
                $\train$: training set, \iid\~sample from $\true$\\
                $z_i$: $i$-th sample in $\train$, $z_i \in \train$\\
                $m$: number of data points in a training set\\
                $K$: number of class\\
                $d$: number of layers (depth) in NN\\
                $h$: number of hidden units (width) in NN\\
                $W^{(l)}$: $l$-th weight matrix\\
                $\param$: parameter of network\\
                $\hyp$: hypothesis, typically depends on $\param$ ($f_\param$)\\
                $\loss{\true}{f}$: expected ($0$-$1$) loss concerning distribution $\true$\\
                $\loss{\sample}{f}$: loss on a data point $\sample$ of a hypothesis $f$\\
                $\loss{\true}{\posterior}$: $\expect{\hyp\sim \posterior}{\loss{\true}{\hyp}}$\\
                $\KL{\posterior}{\prior}$: KL divergence\\
                $\nabla_\param$: derivative concerning parameter $\param$\\
                $\nabla^2_\param$: Hessian concerning parameter $\param$\\
                $\frobenius{W}$: Frobenius norm of a matrix $W$\\
                $\hyp(\sample)$: output of a hypothesis $\hyp$ at a data point $\sample$\\
                $\vec{x}[i]$: the $i$-th element of a vector $\vec{x}$\\
                $A[i,j]$: the $(i,j)$-th element of a matrix $A$\\
                $y_\sample$: label of data point $\sample$\\
                $\idmat$: identity matrix\\
                \bottomrule
            \end{tabular}
        \end{center}
    \end{table}

  \section{Flat minima from PAC-Bayesian perspective}
  \label{sec:flat-minima-pac-bayes}

    In this section, we introduce a fundamental PAC-Bayesian generalization error bound and its connection to flat minima provided by prior work.
    Table\,\ref{table:notation} summarizes notations used in this paper.

    \subsection{PAC-Bayesian generalization error bound}
    \label{sec:flat-minima-pac-bayes:bound}

      One of the most basic PAC-Bayesian generalization error bounds is the following\,\citep{PACBayesInference,PropertyVarApp}.

      For any distribution $\true$, any set $\mathcal{H}$ of classifiers, any distribution $\prior$ of support $\mathcal{H}$, any $\delta\in (0,1]$ and any nonnegative real number $\lambda$, we have, with probability at least $1-\delta$,
      \begin{align}
        \label{eq:pac-bayes}
        &\loss{\true}{\posterior} \leq
        \loss{\train}{\posterior} + \lambda^{-1}\left(\KL{\posterior}{\prior} - \ln\delta + \Psi(\lambda, m)\right),
      \end{align}
      where
      \begin{equation}
        \Psi(\lambda,m) := \ln \expect{\hyp\sim\prior, \train'\sim\true^m}{\exp\left(\lambda(\loss{\true}{\hyp} - \loss{\train'}{\hyp})\right)}.
      \end{equation}
      
      We can provide calculable bounds on $\Psi(\lambda,m)$ depending on which types of loss functions we use\,\citep{PACBayesInference}.
      Especially when we use the $0$-$1$ loss, $\Psi(\lambda, m)$ can be bounded by
\ificml
      \begin{equation}
        \frac{\lambda^2}{2m}.
      \end{equation}
\else
        $\lambda^2 / 2m$.
\fi
      Note that this does not depend on the choice of priors.
      In this paper, we mainly treat the \zoloss.

      We reorganize the PAC-Bayesian bound\,\eqref{eq:pac-bayes} for later use as follows.
      \begin{align}
        \loss{\true}{\hyp}
        =& \underbrace{\loss{\true}{\hyp} - \loss{\true}{\posterior}}_{\text{(A)}} + \loss{\true}{\posterior}\nonumber\\
        \leq& \text{(A)} +
        \loss{\train}{\posterior}\nonumber\\
        &+ \underbrace{\lambda^{-1}\KL{\posterior}{\prior}}_{\text{(B)}} + \lambda^{-1}\left(\ln\frac{1}{\delta} + \Psi(\lambda, m)\right)\nonumber\\
        =& \text{(A)} + \text{(B)} +
        \underbrace{\loss{\train}{\posterior} - \loss{\train}{\hyp}}_{\text{(C)}} + \loss{\train}{\hyp}\nonumber\\
        \label{eq:pac-bayes-D}
        &+\lambda^{-1}\left(\ln\frac{1}{\delta} + \Psi(\lambda, m)\right).
      \end{align}
      Similar decompositions can be found in prior work\,\citep{Nonvacuous,ExploreGeneralization,PACBayesianSpectralMargin}.
      We use a different PAC-Bayes bound\,\eqref{eq:pac-bayes} for later analysis, but they are essentially the same.\footnote{To apply our analysis,
        we can also use some other PAC-Bayesian bounds such as Theorem 1.2.6 in \citet{PACBayesSupervisedClassification},
        which is known to be relatively tight in some cases and successfully provided empirically nontrivial bounds for ImageNet scale networks in \citet{NonvacuousImageNet}.
      }
      The original PAC-Bayesian bound\,\eqref{eq:pac-bayes} is for a stochastic classifier $\posterior$, but the reorganized one\,\eqref{eq:pac-bayes-D} is a bound for a deterministic classifier $\hyp$.

    \subsection{PAC-Bayesian view of flat minima}
    \label{sec:flat-minima-pac-bayes:view}

      Flat minima, which are the noise stability of the training loss with respect to parameters, naturally correspond to (C) in Eq.\,\eqref{eq:pac-bayes-D}.
      When (C) is sufficiently small, we can expect that (A) in \,\eqref{eq:pac-bayes-D} is also small.
      Similarly to existing work\,\citep{NotBounding,FlatMinima,Nonvacuous,CompressionBound}, we focus on analyzing terms (B) and (C).

    \subsection{Effect of noises under second-order approximation}
    \label{sec:flat-minima-pac-bayes:soa}

      To connect PAC-Bayes analysis with the Hessian of the loss landscape as prior work\,\citep{LargeBatchTraining,SharpMinimaGeneralize,HessianAnalysis},
      we consider the second-order approximation of some surrogate loss functions.
      We use the unit-variance Gaussian as the posterior of parameters.
      Then the term (C) in the PAC-Bayesian bound\,\eqref{eq:pac-bayes-D} can be calculated as
      \begin{align}
        \loss{\train}{\posterior} - \loss{\train}{\hyp_\param}
        &= \expect{\vec{\epsilon} \sim \mathcal{N}(\param, \vec{I})}{\loss{\train}{\hyp_{\param + \vec{\epsilon}}}} - \loss{\train}{\hyp_\param} \nonumber\\
        &\approx \expect{\vec{\epsilon}\sim\mathcal{N}(\vec{0},\vec{I})}{\vec{\epsilon}^\top \nabla_\param^2\loss{\train}{\hyp_\param} \vec{\epsilon}}\nonumber\\
        &= \mathrm{Tr}\left(\nabla_\param^2\loss{\train}{\hyp_\param}\right)\nonumber\\
        \label{eq:trace-hessian}
        &= \sum_i \nabla_{\param_i}^2\loss{\train}{\hyp_\param}.
      \end{align}
      Thus, we can approximate the term (C) by the trace of the Hessian.
      % TODO(tsuzuku) prior workでも話されていたことを付け加える
      There are two issues in using the trace of the Hessian as a sharpness metric.
      First, we used unit-variance Gaussians for all parameters, which might not necessarily be the best choice.
      Second, we ignored the effect of the KL-divergence term\,(B).
      While prior work already pointed out these issues\,\citep{Nonvacuous,ExploreGeneralization}, there have not been methods to analyze the two terms jointly.
      The two flaws are the keys of our analysis described in the next section.

  \section{Warm-up: Matrix-normalized flat minima}
  \label{sec:matrix-normalized-flat-minima}

    In this section, we modify flat minima to make them invariant to the transformations proposed in \citet{SharpMinimaGeneralize}.
    An example of networks we consider is the following network with one hidden layer.
    \begin{equation}
      \hyp_\param(x) = W^{(2)}(\mathrm{ReLU}(W^{(1)}(x))).
    \end{equation}
    Weight matrices $W^{(1)}$ and $W^{(2)}$ are subsets of the parameters $\param$.
    Prior work has used some unit-variance Gaussian for the prior and the posterior as discussed in \Sec{sec:flat-minima-pac-bayes:soa}.
    This corresponds to the assumption that all parameters have the same scale.
    However, we already know that the scale can vary from weight matrix to weight matrix.
    In other words, the current choice of priors does not well capture our prior knowledge.
    To cope with this problem, we explicitly make parameters' priors have uncertainty in their scale.
    This section implies that we need to multiply the scale of the loss landscape by the scale of parameters.\footnote{Our
      loss landscape rescaling is slightly different from simply multiplying the scale of parameters to the Hessian\,\eqref{eq:matrix-normalized-flat-minima}.
      Its benefits are discussed in \Sec{sec:connection:fisher-rao}.
    }

    \subsection{Controlling prior variance}
    \label{sec:matrix-normalized-flat-minima:kl}

      In this subsection, we first revisit the technique to control the variance parameters of the Gaussian priors after training\,\citep{KeepNNSimple}.
      Next, we consider its effect on the KL divergence term\,(B) in \eqref{eq:pac-bayes-D}.
      Following standard practice, we use a Gaussian with zero-mean and diagonal covariance matrix ($\sigma_\prior^2\idmat$) where $\sigma_\prior > 0$.
      We also use a Gaussian with mean $\param$ and covariance $\sigma_\posterior^2I$, $\sigma_\posterior > 0$ as the posterior.
      The mean $\param$ is the parameters of the network.

      We first show how to control the variance parameters of the Gaussian priors after training.
      We introduce a hyperprior to the standard deviation $\sigma_\prior$ of parameters per weight matrix\footnote{We
        set other hyperpriors to bias terms similarly. Applying the discussion concerning weight matrices to bias terms is straightforward and thus omitted.
      }\footnote{We
        can also use union bound arguments as alternatives\,\citep{NotBounding,Nonvacuous,PACBayesianSpectralMargin}.
      }\,\citep{KeepNNSimple}.
      To make the prior variance invariant to rescalings, we need to use some special hyperpriors.
      As the hyperprior, we use a uniform prior over a finite set of real numbers.\footnote{Alternatively,
        we can use log-uniform prior over finite domain to make the prior variance invariant to rescalings.
      }
      Especially, we use a set of positive numbers representable by a floating point number as the hyperprior.

      To see how our hyperprior removes scale dependence from the PAC-Bayesian bound\,\eqref{eq:pac-bayes}, we consider its effect to the KL term (B) in \eqref{eq:pac-bayes-D}.
      We can write the KL term as follows.
      \begin{equation}
      \label{eq:pac-bayes:kl}
        \KL{\posterior}{\prior} =
        \sum_{l}\ln \frac{\sigma^{(l)}_\prior}{\sigma^{(l)}_\posterior} + \frac{\frobenius{W^{(l)}}^2 + (\sigma^{(l)}_\posterior)^2}{2(\sigma^{(l)}_\prior)^2} + \const,
      \end{equation}
      where $W^{(l)}$ is the $l$-th weight matrix and $(\sigma^{(l)}_\prior)^2$ and $(\sigma^{(l)}_\posterior)^2$ are the prior variance and the posterior variance of the $i$-th weight matrix, respectively.
      When we fix the prior variances, the KL term is proportional to the squared Frobenius norm of parameters.
      However, since we introduced the special prior, we can arbitrarily change the prior variance after training and control the KL term.
      To minimize the KL divergence term, the prior variance $(\sigma^{(l)}_\prior)^2$ is set to the same value as the posterior variance $(\sigma^{(l)}_\posterior)^2$.
      Below, we use $(\sigma^{(l)})^2$ to denote both $(\sigma^{(l)}_\prior)^2$ and $(\sigma^{(l)}_\posterior)^2$.
      Now, thanks to our hyperprior, we can write the KL divergence term as
      \begin{equation}
      \label{eq:kl-after-reparam}
        \sum_{i}\frac{\frobenius{W^{(l)}}^2}{2(\sigma^{(l)})^2} + \const,
      \end{equation}
      where $\sigma^{(l)}$ are parameters we can tune after training.
      The constant only depends on the number of weight matrices, which is much smaller than the total number of parameters.
      The KL divergence term\,\eqref{eq:kl-after-reparam} has additional flexibility to deal with the scale of weight matrices because we can scale $\sigma^{(l)}$ after training.

    \subsection{Defining matrix-normalized flat minima}
    \label{sec:matrix-normalized-flat-minima:define}

      In this subsection, we show how to tune the variances $(\sigma^{(l)})^2$ introduced in \Sec{sec:matrix-normalized-flat-minima:kl}.
      Deciding the value of the variances induces our definition of scale-invariant flat minima.
      To minimize the PAC-Bayesian bound\,\eqref{eq:pac-bayes}, we choose the variance to minimize the following quantity.
      \begin{equation}
        \label{eq:weight-target}
        \loss{\train}{\posterior} - \loss{\train}{\hyp} + \lambda^{-1}\KL{\posterior}{\prior}.
      \end{equation}
      % TODO(tsuzuku) modelという動詞は適切か?
      First, we model the loss function $\lossfun{\train}$ by second-order approximation.\footnote{The
        choice of the surrogate loss function to calculate the Hessian is discussed in \Sec{sec:matrix-normalized-flat-minima:surrogate-loss}.
      }
      For the sake of notational simplicity, we introduce the following quantity for each weight matrix $W^{(l)}$.
      \begin{equation}
        \label{eq:weight-hessian}
        H^{(l)} := \sum_{i, j} \frac{\partial^2\loss{\train}{\hyp}}{\partial W^{(l)}[i,j]\partial W^{(l)}[i,j]},
      \end{equation}
      where $w$ are parameters in the weight matrix.
      % TODO(tsuzuku) for or concerning
      The quantity $H^{(l)}$ is the sum of the diagonal elements of the Hessian of the training loss function for the weight matrix $W^{(l)}$.
      Now, the quantity\,\eqref{eq:weight-target} can be approximated by
      \begin{equation}
        \label{eq:weight-target-second}
        \eqref{eq:weight-target} \approx \sum_{l} \left( H^{(l)}(\sigma^{(l)})^2 + \frac{\frobenius{W^{(l)}}^2}{2\lambda(\sigma^{(l)})^2}\right) + \const,
      \end{equation}
      where $\left(\sigma^{(l)}\right)^2$ is the variance associated to the weight matrix $W^{(l)}$.
      With an assumption that the Hessian is positive semidefinite,
      the quantity\,\eqref{eq:weight-target-second} is minimized when we set
      \begin{equation}
        (\sigma^{(l)})^2 = \sqrt{\frac{\frobenius{W^{(l)}}^2}{2\lambda H^{(l)}}}.
      \end{equation}
      By inserting this to the quantity\,\eqref{eq:weight-target-second}, we get
      \begin{equation}
        \label{eq:matrix-normalized-flat-minima-all}
        \eqref{eq:weight-target} \approx \sqrt{\frac{2}{\lambda}}\sum_l\sqrt{\frobenius{W^{(l)}}^2H^{(l)}} + \const
      \end{equation}
      No matter what $\lambda$ achieves the infimum in the PAC-Bayes bound\,\eqref{eq:pac-bayes}, the smaller the quantity\,\eqref{eq:matrix-normalized-flat-minima-all} is, the smaller the bound\,\eqref{eq:pac-bayes} is.
      Thus, we can use the following as a measure of generalization:
      \begin{equation}
        \label{eq:matrix-normalized-flat-minima}
        \sum_l\sqrt{\frobenius{W^{(l)}}^2H^{(l)}}.
      \end{equation}
      We refer to this quantity as matrix-normalized sharpness.
      Intuitively, we scale the sharpness by the scale of each weight matrix.
      % TODO(tsuzuku) appendixとかで示しておく
      Matrix-normalized sharpness\,\eqref{eq:matrix-normalized-flat-minima} is invariant to the rescaling of parameters proposed in \citet{SharpMinimaGeneralize}.
      Thus, we can overcome one of the open problems by considering the effect of both terms (B) and (C) in PAC-Bayesian bound\,\eqref{eq:pac-bayes-D}.
      Its connection to minimum description length arguments, which are the basis of flat minima arguments, is discussed in \Sec{sec:connection:mdl}.

  \section{Normalized flat minima}
  \label{sec:normalized-flat-minima}

    In this section, we point out a scale dependence of matrix-wise capacity control which is similar to the prior flat minima definitions.
    To remove the scale dependence, we extend matrix-normalized flat minima and define normalized flat minima.
    The new definition provides improved invariance while enjoying a reduced effective number of parameters in the constant term for a better generalization guarantee.
    The extension has a more complicated form than matrix-normalized flat minima.
    However, it is similar in a sense that it multiplies the scale of parameters to the loss curvatures.

    \subsection{Scale dependence of matrix-wise capacity control}
    \label{sec:normalized-flat-minima:insufficiency}

      First, we propose a transformation different from \citet{SharpMinimaGeneralize} that changes the scale of the Hessian of networks arbitrarily.
      Let us consider a simple network with a single hidden layer and ReLU activation.
      We denote this network as
      \begin{equation}
        \hyp_\param(x) = W^{(2)}(\mathrm{ReLU}(W^{(1)}(x))).
      \end{equation}
      We can scale the $i$-th column of $W^{(2)}$ by $\alpha > 0$ and $i$-th row of $W^{(1)}$ by $1 / \alpha$ without modifying the function that the network represents as follows.\footnote{Running
        examples of the transformation can be found in \Appendix{sec:appendix:running-examples:row-column-scaling}.
      }
      \begin{align}
        W^{(1)}[i,:] &\leftarrow \frac{W^{(1)}[i,:]}{\alpha}, \\
        W^{(2)}[:,i] &\leftarrow \alpha W^{(2)}[:,i].
      \end{align}
      Since we are using the ReLU activation function, which has positive homogeneity, this transformation does not change the represented function.
      By the transformation, the scale of the diagonal elements of the Hessian corresponding to the $i$-th row of $W^{(1)}$ are scaled by $\alpha^2$.
      This can cause essentially the same effect with the transformation proposed by \citet{SharpMinimaGeneralize}.

      The transformation reveals a scale dependence of matrix-norm based generalization error bounds as follows.
      Assume $W^{(1)}$ has at least two non-zero rows and $W^{(2)}$ has at least two non-zero columns.
      Using the transformation, we can make both $W^{(1)}$, and $W^{(2)}$ have at least one arbitrarily large element.
      In other words, both weight matrices have arbitrarily large spectral norms and Frobenius norms.
      Also, the stable rank of the two matrices become arbitrarily close to one.
      Thus, the matrix-norm based capacity control\,\citep{SpectralMargin,PACBayesianSpectralMargin} suffers from the same scale dependence as the prior definitions of flatness.

    \subsection{Improving invariance of flatness}
    \label{sec:normalized-flat-minima:invariance}

      To address the newly revealed scale dependence in \Sec{sec:normalized-flat-minima:insufficiency}, we modify the choice of the hyperprior discussed in \Sec{sec:matrix-normalized-flat-minima:define}.
      We introduce a parameter $\sigma_i$ for the $i$-th row and $\sigma'_j$ for the $j$-th column and use the product of them as variance.\footnote{In
        some parameters such as bias terms and scaling parameters in normalization layers,
        setting variance parameters per row corresponds to applying naive parameter counting for these parameters.
        Thus, noise induced by the posterior become $0$ and the KL-term become a constant that scales with the number of such parameters.
      }
      In other words, we set the variance of $W_{i,j}$ to $\sigma_i\sigma'_j$.
      For the priors of $\sigma$ and $\sigma'$, we use the same priors with \Sec{sec:matrix-normalized-flat-minima}.
      Setting the variances per row and column makes the constant term in the KL-term $O(hd)$.
      This is still much smaller than setting variance per parameter, which scales $O(h^2d)$.    
      Applying the same discussion as \Sec{sec:matrix-normalized-flat-minima:define}, we define normalized sharpness as the sum of the solutions of the following optimization problem defined for each weight matrix.
      \begin{equation}
        \label{eq:normalized-flat-minima}
        \min_{\vec{\sigma},\vec{\sigma'}} \sum_{i,j}\left( \frac{\partial^2\loss{\train}{\hyp}}{\partial W[i,j]\partial W[i,j]}(\sigma_i\sigma'_j)^2 + \frac{W[i,j]^2}{2\lambda(\sigma_i\sigma'_j)^2}\right).
      \end{equation}

      In convolutional layers, since the same filter has the same scale, we only need to set the hyperprior on the input and output channels.
      When $\lambda$ changes to $\lambda'$, the normalized sharpness\,\eqref{eq:normalized-flat-minima} is scaled by $\sqrt{\lambda/\lambda'}$
      as matrix-normalized flat minima\,(\Sec{sec:matrix-normalized-flat-minima}).
      Thus, networks with smaller normalized flatness at some choice of $\lambda$, e.g., $1/2$, also have smaller normalized sharpness at other choices of $\lambda$.
      Below, we set $\lambda = 1/2$ for simpler calculation.

    \subsection{Practical calculation}
    \label{sec:normalized-flat-minima:calculation}

      We present a practical calculation technique to solve the optimization problem\,\eqref{eq:normalized-flat-minima}.
      First, we reparametrize variance parameters $\sigma_i$ and $\sigma'_j$ as follows.
      \begin{align}
        e^{\alpha_i} &:= \sigma_i^2, \\
        e^{\beta_j} &:= \sigma_j'^2.
      \end{align}
      Fortunately, the optimization problem\,\eqref{eq:normalized-flat-minima} is convex with respect to $\alpha_i$ and $\beta_i$.\footnote{See
        \Appendix{sec:appendix:convexity}.
      }
      It is straightforward to see that the convexity also holds with convolutional layers.
      Thus, we can estimate the near optimal value of $\sigma_i$ and $\sigma'_j$ by gradient descent.
      Details of the gradient calculation can be found in \Appendix{sec:appendix:gd}.
      Figure\,\ref{alg:ns} shows the pseudo code of the normalized sharpness calculation.

      \begin{figure}[t]
        \small
        \begin{center}
          \begin{algorithm}[H]
            \caption{Normalized sharpness calculation}
            \begin{algorithmic}
              \STATE // Calculate diagonal elements of the Hessian
              \STATE
              $\vec{h} = \expect{\vec{\epsilon}\sim\mathcal{N}(\vec{0},\vec{1})}{\vec{\epsilon}\odot\frac{\nabla_{\param} \loss{\train}{\hyp_{\param + r\vec{\epsilon}}}
              - \nabla_{\param} \loss{\train}{\hyp_{\param - r\vec{\epsilon}}}}{2r}}$
              \STATE // Positive semi-definite assumption might be false
              \STATE $\vec{h} \leftarrow \mathrm{Clip}(\vec{h}, \vec{0})$
              \STATE $S \leftarrow 0$ // normalized sharpness
              \STATE // Solve \eqref{eq:normalized-flat-minima} for each weight matrix
              \FOR{\textbf{each} weight matrix $W^{(l)}$}
                \WHILE{not converged}
                  \STATE // reparametrization to make the loss convex
                  \STATE $\vec{\sigma} \leftarrow \exp\left(\vec{\alpha}\right)$
                  \STATE $\vec{\sigma}' \leftarrow \exp\left(\vec{\beta}\right)$
                  \STATE $S^{l} \leftarrow 0$ // normalized sharpness of $W^{(l)}$
                  \FOR{\textbf{each} row $i$}
                    \FOR{\textbf{each} column $j$}
                      \STATE // $\vec{h}^{(l)}[i,j]$ is an element of $\vec{h}$
                      \STATE // corresponding to $W^{(l)}[i,j]$
                      \STATE $S^{l} \leftarrow S^{l} + \vec{h}^{(l)}[i,j](\vec{\sigma}[i]\vec{\sigma'}[j])^2$
                      \STATE $S^{l} \leftarrow S^{l} + W^{(l)}[i,j]^2 / (\vec{\sigma}[i]\vec{\sigma'}[j])^2$
                    \ENDFOR
                  \ENDFOR
                  \STATE $\mathrm{SGDUpdate}(L, \vec{\alpha}, \vec{\beta})$
                \ENDWHILE
                \STATE $S \leftarrow S + S^{(l)}$
              \ENDFOR
              \STATE $\mathrm{return}$ $S$
            \end{algorithmic}
          \end{algorithm}
        \end{center}
        \caption{
          Calculation of the normlaized sharpness\,\eqref{eq:normalized-flat-minima}.
        }
        \label{alg:ns}
      \end{figure}

    \subsection{Choice of the surrogate loss function}
    \label{sec:matrix-normalized-flat-minima:surrogate-loss}

      When we measure the generalization gap using the \zoloss, which is not differentiable with respect to parameters, we need to use surrogate loss functions.
      The choice of the surrogate loss functions needs special care when we use flatness for model comparison.
      For the comparison to make sense, the value of the normalized sharpness is preferable not to change when the accuracy of the models does not change.
      Thus, the surrogate loss function is better to make the normalized sharpness invariant against some changes that do not change accuracy such as scalings and shifting of the networks' outputs.
      For example, the cross-entropy loss taken after softmax does not satisfy the first condition.
      Thus, using the loss function makes the model comparison less meaningful.
      While the above conditions do not make the choices of the surrogate loss function unique, we heuristically use the following loss.
      \begin{align}
      \label{eq:nsce}
        -\ln\left(\frac{\exp\left(f'(\sample)[y_\sample]\right)}{\sum_i \exp\left(f'(\sample)[i]\right)}\right),
      \end{align}
      where
      \begin{align}
        &\mu := \frac{1}{K}\sum_i f(\sample)[i],\\
        &f'(\sample) := \frac{f(\sample)}{\sqrt{\frac{1}{K}\sum_i\left(f(\sample)[i] - \mu\right)^2}},\\
      \end{align}
      $\hyp(\sample)$ is an output of a network, $y_\sample$ is a label of $\sample$, and $K$ is the number of classes.
      We refer to the loss function as normalized-softmax-cross-entropy loss.
      We use this loss function in later experiments\,(\Sec{sec:experiments}).

  \section{Numerical evaluations}
  \label{sec:experiments}

    We numerically justify the insights from the previous sections.
    We specifically check the followings.
    \begin{itemize}
      \item Normalized sharpness\,(\Sec{sec:normalized-flat-minima}) distinguishes models trained on random labels\,(\Sec{sec:experiments:nf-rl}).
      \item Scale dependence of existing sharpness metric can be harmful in common settings, not only artificial ones\,(\Sec{sec:experiments:f-rl}).
      \item Normalized sharpness better captures generalization than existing sharpness metrics\,(\Sec{sec:experiments:f-rl}).
    \end{itemize}
    Detailed experimental setups are described in \Appendix{sec:appendix:setup}.

    \subsection{Distinguishing models trained on random labels}
    \label{sec:experiments:nf-rl}

      We checked whether normalized sharpness can distinguish models trained on random labels.
      Hypotheses which fit random labels belong to hypothesis classes such that Rademacher complexity is $1$.
      Thus, if normalized sharpness captures generalization reasonably well, it should have a larger value for networks trained on random labels.

      \paragraph{Quantities:}
        We investigated the correlation between normalized sharpness\,\eqref{eq:normalized-flat-minima} and generalization gap defined by (train accuracy $-$ test accuracy).
        Sharper minima are expected to have larger generalization gaps.
        A loss function\,\eqref{eq:nsce} was used as the surrogate loss function of the \zoloss.

      \paragraph{Set up:}
        We trained a multilayer perceptron with three hidden layers and LeNet\,\citep{LeNet} on MNIST\,\citep{MNIST} and LeNet and Wide ResNet\,\citep{WideResNet} with 16 layers and width factor $4$ on CIFAR-10\,\citep{CIFAR} for 100 times for each pair.
        At each run, we randomly selected the ratio of random labels from $0$ to $1$ at $0.1$ intervals.
        We used Adam optimizer and applied no regularization or data augmentation so that the training accuracy reached near $1$ even with random labels.

\begin{figure}[t]
          \begin{minipage}{.49\hsize}
            \begin{center}
              \includegraphics[width=\linewidth]{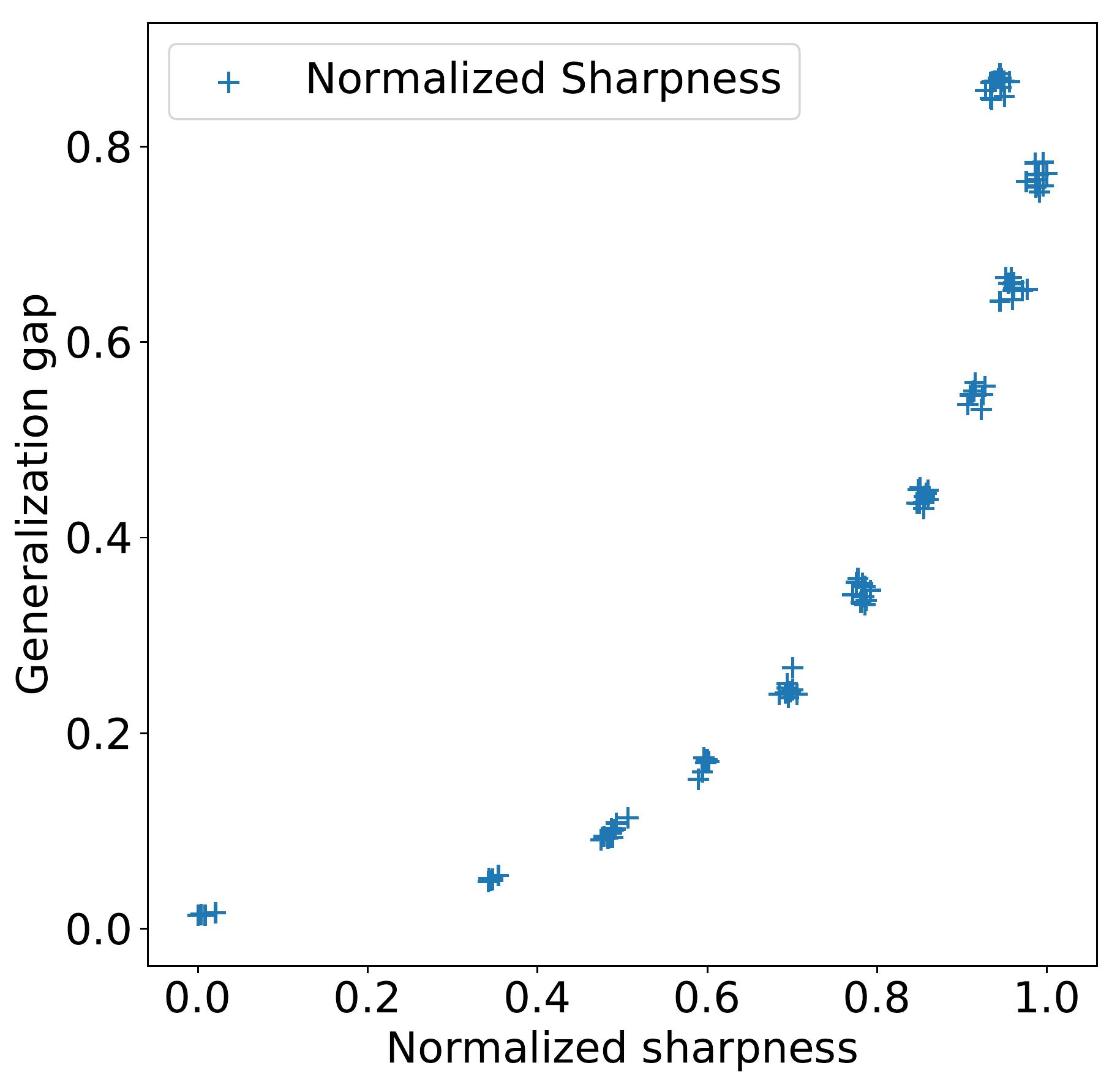}
            \end{center}
          \end{minipage}
          \begin{minipage}{.49\hsize}
            \begin{center}
              \includegraphics[width=\linewidth]{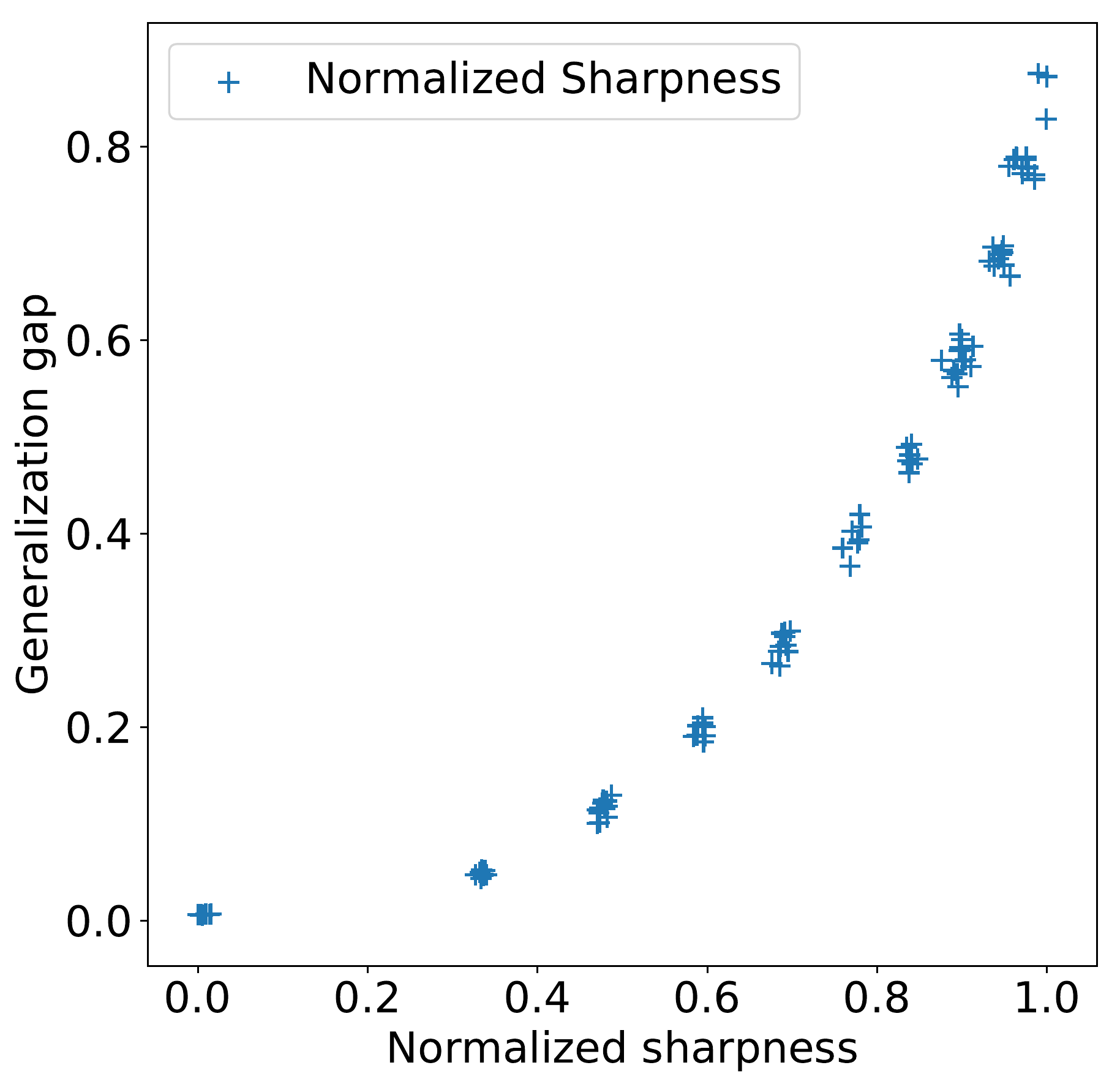}
            \end{center}
          \end{minipage}
          \begin{minipage}{.49\hsize}
            \begin{center}
              \includegraphics[width=\linewidth]{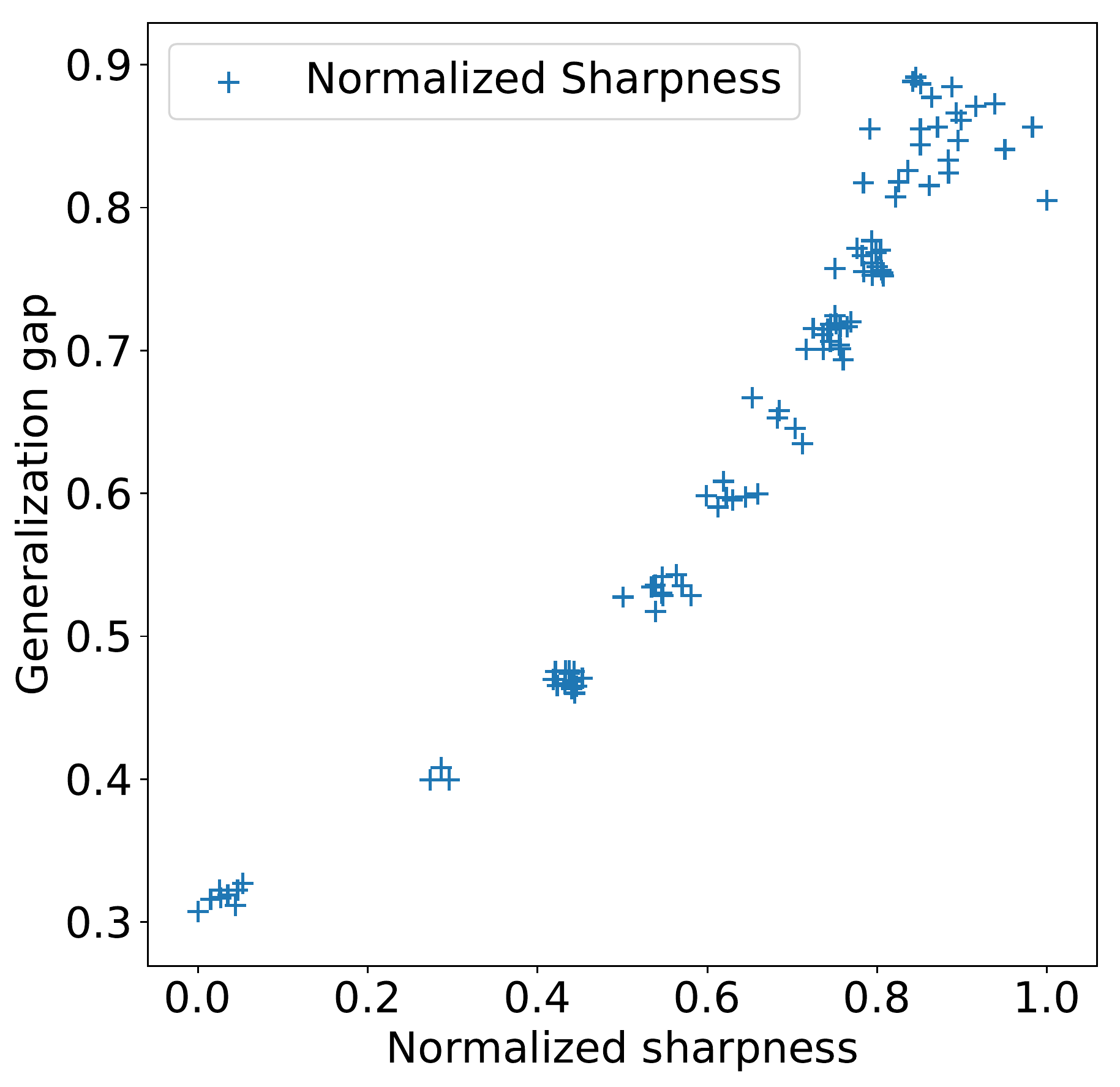}
            \end{center}
          \end{minipage}
          \begin{minipage}{.49\hsize}
            \begin{center}
              \includegraphics[width=\linewidth]{img/correlation_result_config_201810_random_label4_cifar10_wres_nf_nsce_0-1_8_normalized.pdf}
            \end{center}
          \end{minipage}
          \caption{
            Scatter plot between normalized sharpness\,\eqref{eq:normalized-flat-minima} and accuracy gap.
            Normalized sharpnesses were rescaled to $[0, 1]$ by their maximum and minimum among the trained networks.
            Top left figure shows the results for multi layer perceptrons on MNIST.
            Top right shows LeNet on MNIST.
            Bottom left shows LeNet on CIFAR-10.
            And Bottom right shows Wide ResNet on CIAFR-10.
          }
          \label{fig:compare-ns-rl}
        \end{figure}

      \paragraph{Results:}
        Figure\,\ref{fig:compare-ns-rl} shows scatter plots of normalized sharpness v.s. accuracy gap for networks trained on MNIST and CIFAR-10.
        The results show that networks tended to have larger normalized sharpness to fit random labels.
        Thus, we can say that normalized sharpness provides reasonably good hierarchy in hypothesis class.
        These results support our analysis concerning normalized flat minima in \Sec{sec:matrix-normalized-flat-minima} and \Sec{sec:normalized-flat-minima}.

    \subsection{Effect of normalization}
    \label{sec:experiments:f-rl}

      We tested how our modification of flat minima change its property.
      We used the same trained model with \Sec{sec:experiments:nf-rl}, but used cross-entropy loss for calculating the Hessian.
      We plotted the trace of the Hessian without normalization\,\eqref{eq:trace-hessian} and the sum of the squared Frobenius norm of the weight matrices\,\eqref{eq:kl-after-reparam}.

\begin{figure}[t]
          \begin{minipage}{.49\hsize}
            \begin{center}
              \includegraphics[width=\linewidth]{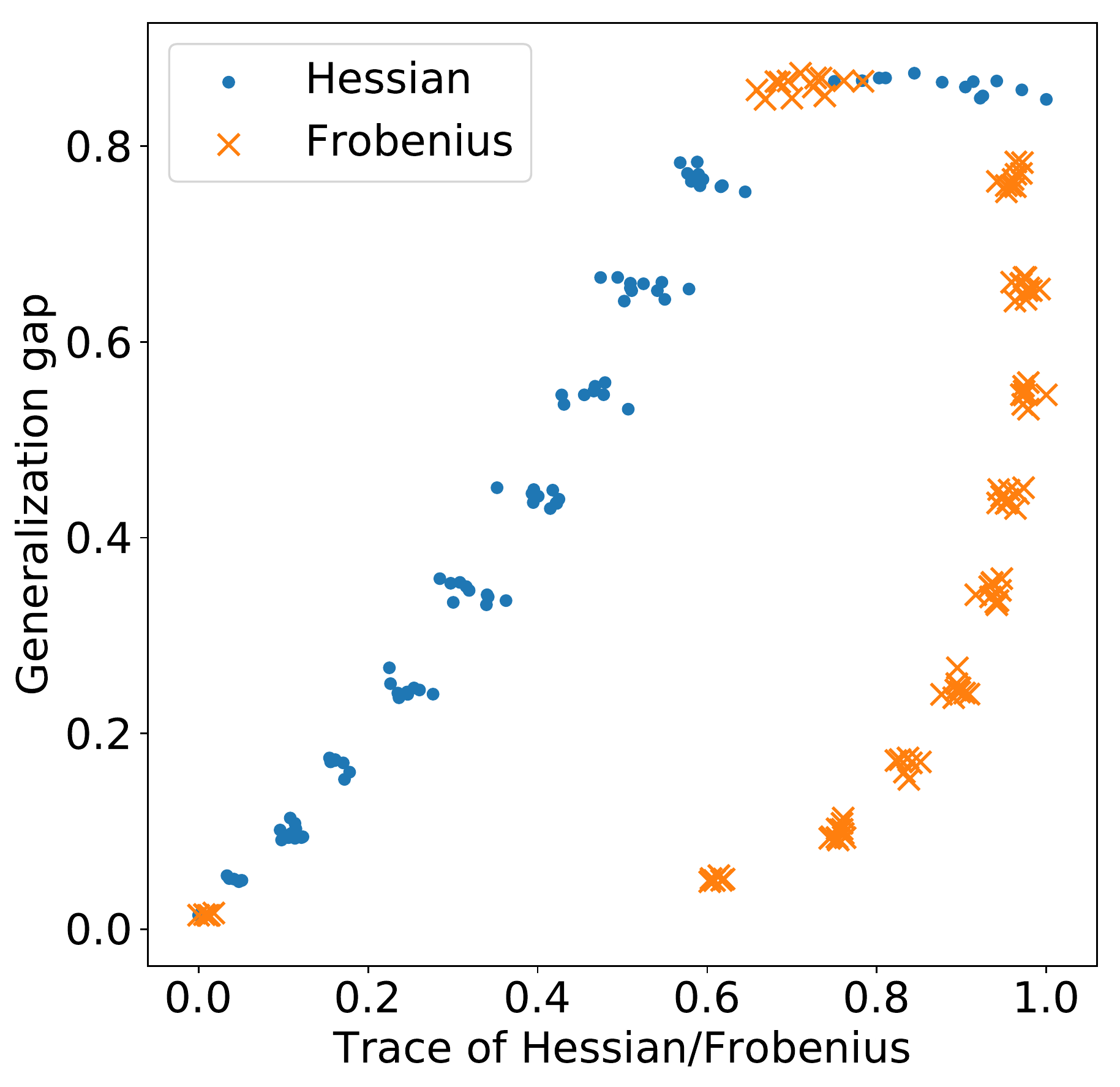}
            \end{center}
          \end{minipage}
          \begin{minipage}{.49\hsize}
            \begin{center}
              \includegraphics[width=\linewidth]{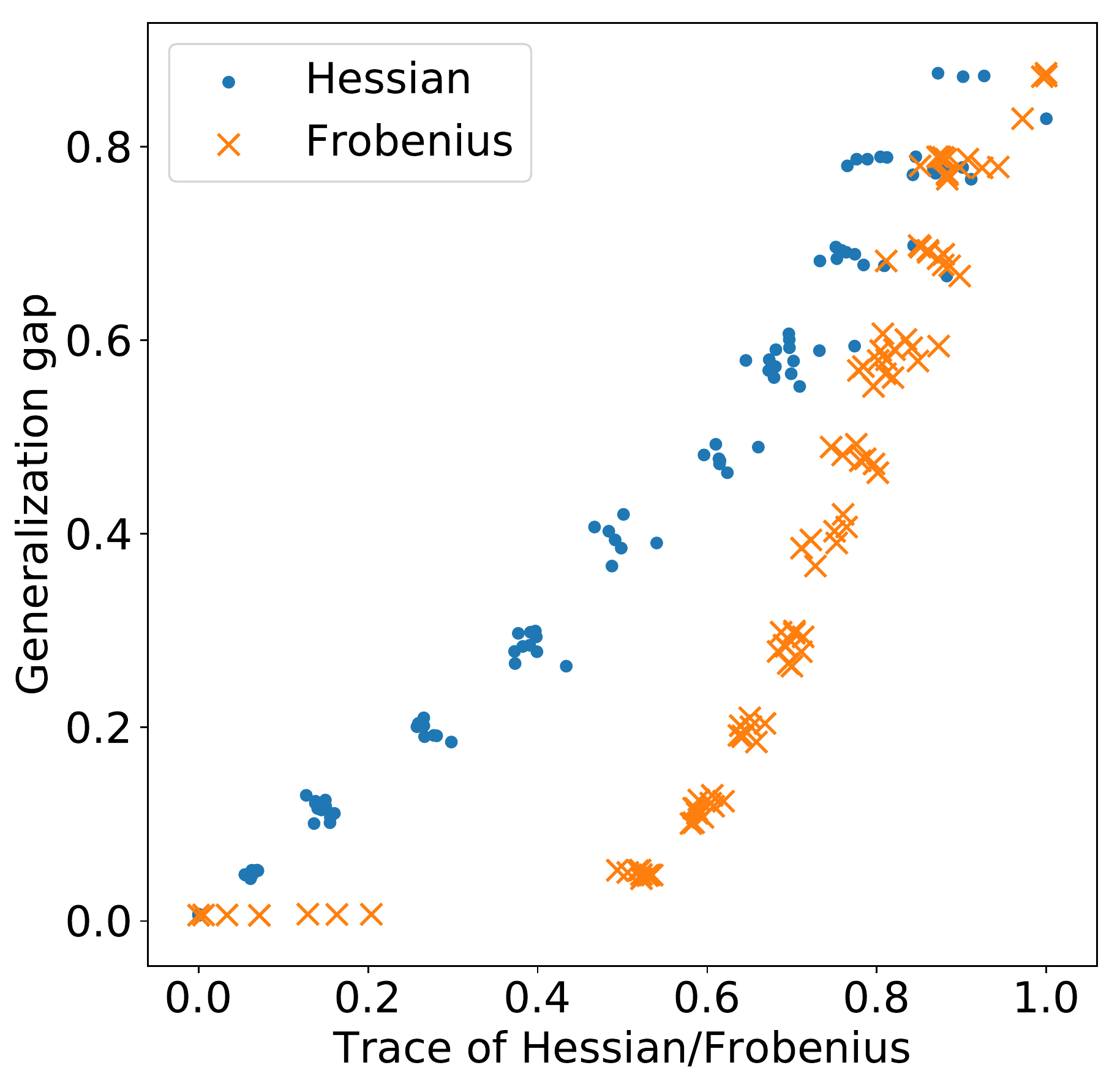}
            \end{center}
          \end{minipage}
          \begin{minipage}{.49\hsize}
            \begin{center}
              \includegraphics[width=\linewidth]{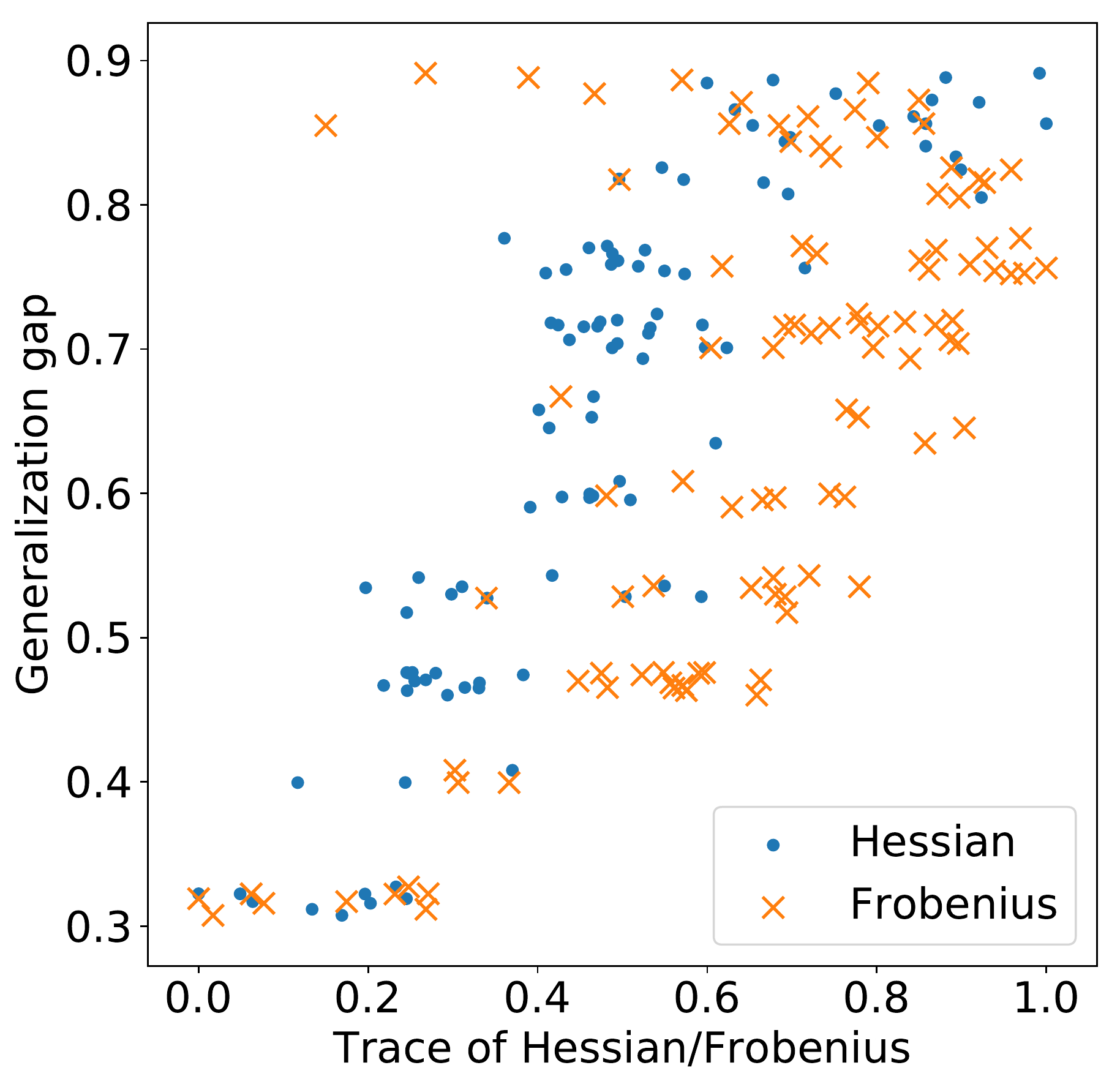}
            \end{center}
          \end{minipage}
          \begin{minipage}{.49\hsize}
            \begin{center}
              \includegraphics[width=\linewidth]{img/correlation_result_config_201810_random_label4_cifar10_wres_f_sce_kl_0-1_8_normalized.pdf}
            \end{center}
          \end{minipage}
          \caption{
            Scatter plot between trace of the Hessian\,\eqref{eq:trace-hessian}, sum of the squared Frobenius norm of weight matrices, and accuracy gap.
            The traces were rescaled to $[0, 1]$ by their maximum and minimum among the trained networks.
            Top left figure shows the results for multi layer perceptrons on MNIST.
            Top right shows LeNet on MNIST.
            Bottom left shows LeNet on CIFAR-10.
            And Bottom right shows Wide ResNet on CIAFR-10.
          }
          \label{fig:compare-s-rl}
        \end{figure}

      Figure\,\ref{fig:compare-s-rl} shows the results.
      Even though sharpness without normalization can also distinguish models trained on random labels to some extent, the signal is weaker compared to normalized sharpness.
      Notably, in larger models with normalization layers, sharpness without normalization lost its ability to distinguish models.
      The result shows that the scaling dependence of the flatness measures can be problematic even in natural settings and also supports the advantages of the normalization.

  \section{Discusion}
  \label{sec:connection}

    In this section, we discuss connections between normalized flat minima and previous studies.

    \subsection{Connection to MDL arguments}
    \label{sec:connection:mdl}

      Minimum description length (MDL)\,\citep{MDL} provides us with a measure of generalization through the amount of the necessary information to describe the model.
      Intuitively, at flat minima, we can use less accurate representations of the parameters and thus requires fewer bits.
      The gain of the description length is quantitatively explained by bits-back arguments\,\citep{KeepNNSimple,BitsBack}.
      According to the theory, we can represent the model by the following number of bits.
      \begin{equation}
      \label{eq:mdl-kl}
        \KL{\posterior}{\prior}.
      \end{equation}
      This is the same with the (B) term in \eqref{eq:pac-bayes-D}.
      Thus, from the minimum description length principle, normalized sharpness balances the effect of the posterior variance and the number of bits we can save.
      For more discussion on the connection to MDL, please refer to \citet{Nonvacuous}.

    \subsection{Comparison with other prior choices}
    \label{sec:connection:local-reparam}

      \citet{VariationalDropout} proposed a local reparametrization trick that removes the scale dependence of the KL term by hyperpriors.
      However, in \citet{VariationalDropout}, reparametrization was performed per parameter.
      This makes the constant in \eqref{eq:kl-after-reparam} scales with the number of parameters.
      Thus, even though the trick removes the scale independence, the resultant KL-term is as good as a naive parameter counting.
      On the other hand, the constant in our analysis scales at most $O(hd)$ compared to their $O(h^2d)$.
      \citet{EmergenceOfInvariance} used the local reparametrization trick to connect information bottleneck\,\citep{InformationBottleneck} with flat minima and PAC-Bayes.
      However, the use of the trick made their discussion vacuous from PAC-Bayesian perspective.
      Reinterpreting information bottleneck using our prior design might provide novel insights.

      \citet{Nonvacuous} and \citet{OverParam} used the initial parameters as the mean of the prior.
      In deep networks, we could not empirically observe its advantages over using zero means.
      Moreover, it provides additional scale dependence to the notion of the flat minima.
      Applying some normalization to the prior mean to make the overall bound scale-invariant might help to utilize the initial state of the networks even in deep networks, but we leave the exploration as future work.

    \subsection{Comparison with Fisher-Rao norm}
    \label{sec:connection:fisher-rao}

      \citet{FisherRaoNorm} proposed the Fisher-Rao norm, which is defined as follows.
      \begin{equation}
        \sum_i \left( \param_i^2 \expect{\sample\sim\train}{\nabla_{\param_i} \loss{\sample}{\hyp_\param}^2} \right).
      \end{equation}
      While the formulation is similar to normalized sharpness, there are three crucial differences.
      First, normalized sharpness uses the Hessian, not Fisher, and directly measures curvature.
      Second, the Fisher-Rao norm is parameter-wise, while the normalized sharpness exploits the parameter structures in neural networks.
      Third, normalized sharpness takes the square root of the Frobenius norms of parameters and the Hessian.
      To highlight an advantage by the third difference, we consider the following network.
      \begin{equation}
        y = \mathrm{ReLU}(W^{(1)}x) + \mathrm{ReLU}(W^{(2)}x).
      \end{equation}
      This type of connection is often found in modern networks\,\citep{ResNext,NASNet}.
      We assume that the following condition is satisfied.
      \begin{equation}
        W^{(1)} = W^{(2)} = W/2.
      \end{equation}
      Next, we rescale the parameters as follows.
      \begin{align}
        &y = \mathrm{ReLU}(W'^{(1)}x) + \mathrm{ReLU}(W'^{(2)}x),\\
        &W'^{(1)} = W,\\
        &W'^{(2)} = \vec{O}.
      \end{align}
      By this rescaling, the Fisher-Rao norm of the weight matrix $W$ becomes half, while the normalized sharpness\,\eqref{eq:normalized-flat-minima} is kept the same.
      The definition of matrix-normalized sharpness\,\eqref{eq:matrix-normalized-flat-minima} is also invariant against this rescaling.
      This additional invariance suggests that our definition better captures generalization.

    \subsection{Supporting empirical findings}
    \label{sec:connection:empirical-findings}

      \citet{VisualizeLoss} showed that we can clearly observe flat/sharp minima when we rescale the loss landscape by the scale of the parameters.
      Especially, they applied normalization filter-wisely rather than layer-wisely in convolutional layers.
      This is closely related to the notion of row and column wise normalization proposed in \Sec{sec:normalized-flat-minima}.
      \citet{AdamW} proposed AdamW and empirically closed generalization gap of trained networks using Adam.
      \citet{ThreeMechanismWeightDecay} connected second order optimization with weight decay such as AdamW with normalized curvature, which plays a key role in this paper.
      Rethinking such optimization methods through the normalized flat minima might be useful to improve them further.

  \section{Conclusion}
  \label{sec:discussion}

    In this paper, we proposed a notion of normalized flat minima, which is free from the known scale dependence.
    The advantages of our definition are as follows.
    \begin{itemize}
      \item It is invariant to transformations from which the prior definitions of flatness suffered.
      \item It can approximate generalization bounds tighter than naive parameter countings.
    \end{itemize}
    Our discussion extends potential applications of the notion of flat minima from the cases when parameters are ``appropriately'' normalized to general cases.
    Experimental results suggest that our analysis is powerful enough to distinguish overfitted models even when models are large and existing flat minima definitions tend to suffer from the scale dependence issues.

    One flaw of the normalized flat minima is that it uses Gaussian for both prior and posterior even though that is standard practice in the literature\,\citep{KeepNNSimple,PACBayesianSpectralMargin}.
    From \citet{NoBarriers} and \citet{AverageWeights}, we know that appropriate posteriors of networks have more complex structures than Gaussians.
    From the minimum description length perspective, using Gaussian limits the compression algorithms of models.
    Recent analyses of compression algorithms for neural networks\,\citep{MDLofDeep} might be useful for better prior and posterior designs.
    \citet{FunctionalBBN} might help to develop methods to define priors and posteriors on function space and calculate the KL-divergence on function space directly.

\ificml
\else
  \section*{Acknowledgement}
    YT was supported by Toyota/Dwango AI scholarship.
    IS was supported by KAKENHI 17H04693.
    MS was supported by the International Research Center for Neurointelligence (WPI-IRCN) at The University of Tokyo Institutes for Advanced Study.
\fi

  \clearpage

  \bibliography{main}

\begin{thebibliography}{43}
\providecommand{\natexlab}[1]{#1}
\providecommand{\url}[1]{\texttt{#1}}
\expandafter\ifx\csname urlstyle\endcsname\relax
  \providecommand{\doi}[1]{doi: #1}\else
  \providecommand{\doi}{doi: \begingroup \urlstyle{rm}\Url}\fi

\bibitem[Achille \& Soatto(2018)Achille and Soatto]{EmergenceOfInvariance}
Achille, A. and Soatto, S.
\newblock {Emergence of Invariance and Disentanglement in Deep
  Representations}.
\newblock \emph{Journal of Machine Learning Research}, 19, 2018.

\bibitem[Alquier et~al.(2016)Alquier, Ridgway, and Chopin]{PropertyVarApp}
Alquier, P., Ridgway, J., and Chopin, N.
\newblock {On the properties of variational approximations of Gibbs
  posteriors}.
\newblock \emph{Journal of Machine Learning Research}, 17\penalty0
  (239):\penalty0 1--41, 2016.

\bibitem[Arora et~al.(2018)Arora, Ge, Neyshabur, and Zhang]{CompressionBound}
Arora, S., Ge, R., Neyshabur, B., and Zhang, Y.
\newblock {Stronger Generalization Bounds for Deep Nets via a Compression
  Approach}.
\newblock In \emph{Proceedings of the 35th International Conference on Machine
  Learning}, volume~80 of \emph{Proceedings of Machine Learning Research}, pp.\
   254--263. PMLR, 10--15 Jul 2018.

\bibitem[Bartlett et~al.(2017)Bartlett, Foster, and Telgarsky]{SpectralMargin}
Bartlett, P.~L., Foster, D.~J., and Telgarsky, M.~J.
\newblock {Spectrally-normalized margin bounds for neural networks}.
\newblock In \emph{Advances in Neural Information Processing Systems 30}, pp.\
  6240--6249. Curran Associates, Inc., 2017.

\bibitem[Blier \& Ollivier(2018)Blier and Ollivier]{MDLofDeep}
Blier, L. and Ollivier, Y.
\newblock {The Description Length of Deep Learning Models Léonard}.
\newblock In \emph{Advances in Neural Information Processing Systems 31}, pp.\
  2220--2230. Curran Associates, Inc., 2018.

\bibitem[Catoni(2007)]{PACBayesSupervisedClassification}
Catoni, O.
\newblock \emph{{Pac-Bayesian Supervised Classification: The Thermodynamics of
  Statistical Learning}}.
\newblock {Institute of Mathematical Statistics}, 2007.

\bibitem[Chaudhari et~al.(2017)Chaudhari, Choromanska, Soatto, LeCun, Baldassi,
  Borgs, Chayes, Sagun, and Zecchina]{EntropySGD}
Chaudhari, P., Choromanska, A., Soatto, S., LeCun, Y., Baldassi, C., Borgs, C.,
  Chayes, J., Sagun, L., and Zecchina, R.
\newblock {Entropy-SGD: Biasing Gradient Descent Into Wide Valleys}.
\newblock In \emph{International Conference on Learning Representations}, 2017.

\bibitem[Dinh et~al.(2017)Dinh, Pascanu, Bengio, and
  Bengio]{SharpMinimaGeneralize}
Dinh, L., Pascanu, R., Bengio, S., and Bengio, Y.
\newblock {Sharp Minima Can Generalize For Deep Nets}.
\newblock In \emph{Proceedings of the 34th International Conference on Machine
  Learning}, volume~70 of \emph{Proceedings of Machine Learning Research}, pp.\
   1019--1028. PMLR, 06--11 Aug 2017.

\bibitem[Draxler et~al.(2018)Draxler, Veschgini, Salmhofer, and
  Hamprecht]{NoBarriers}
Draxler, F., Veschgini, K., Salmhofer, M., and Hamprecht, F.
\newblock Essentially no barriers in neural network energy landscape.
\newblock In \emph{Proceedings of the 35th International Conference on Machine
  Learning}, volume~80 of \emph{Proceedings of Machine Learning Research}, pp.\
   1309--1318. PMLR, 10--15 Jul 2018.

\bibitem[Dziugaite \& Roy(2017)Dziugaite and Roy]{Nonvacuous}
Dziugaite, G.~K. and Roy, D.~M.
\newblock {Computing Nonvacuous Generalization Bounds for Deep (Stochastic)
  Neural Networks with Many More Parameters than Training Data}.
\newblock In \emph{Proceedings of the Thirty-Third Conference on Uncertainty in
  Artificial Intelligence}, 2017.

\bibitem[Germain et~al.(2016)Germain, Bach, Lacoste, and
  Lacoste-Julien]{PACBayesInference}
Germain, P., Bach, F., Lacoste, A., and Lacoste-Julien, S.
\newblock {PAC-Bayesian Theory Meets Bayesian Inference}.
\newblock In \emph{Advances in Neural Information Processing Systems 29}, pp.\
  1884--1892. Curran Associates, Inc., 2016.

\bibitem[Hinton \& van Camp(1993)Hinton and van Camp]{KeepNNSimple}
Hinton, G.~E. and van Camp, D.
\newblock {Keeping the Neural Networks Simple by Minimizing the Description
  Length of the Weights}.
\newblock In \emph{Proceedings of the Sixth Annual Conference on Computational
  Learning Theory}, COLT '93, pp.\  5--13. ACM, 1993.
\newblock ISBN 0-89791-611-5.

\bibitem[Hochreiter \& Schmidhuber(1997)Hochreiter and Schmidhuber]{FlatMinima}
Hochreiter, S. and Schmidhuber, J.
\newblock {Flat Minima}.
\newblock \emph{Neural Computation}, 9\penalty0 (1):\penalty0 1--42, 1997.

\bibitem[Hoffer et~al.(2017)Hoffer, Hubara, and Soudry]{TrainLonger}
Hoffer, E., Hubara, I., and Soudry, D.
\newblock {Train longer, generalize better: closing the generalization gap in
  large batch training of neural networks}.
\newblock In \emph{Advances in Neural Information Processing Systems 30}, pp.\
  1731--1741. Curran Associates, Inc., 2017.

\bibitem[Honkela \& Valpola(2004)Honkela and Valpola]{BitsBack}
Honkela, A. and Valpola, H.
\newblock {Variational Learning and Bits-back Coding: An Information-theoretic
  View to Bayesian Learning}.
\newblock \emph{Trans. Neur. Netw.}, 15\penalty0 (4):\penalty0 800--810, July
  2004.
\newblock ISSN 1045-9227.

\bibitem[Ioffe \& Szegedy(2015)Ioffe and Szegedy]{BatchNorm}
Ioffe, S. and Szegedy, C.
\newblock {Batch Normalization: Accelerating Deep Network Training by Reducing
  Internal Covariate Shift}.
\newblock In \emph{Proceedings of the 32nd International Conference on Machine
  Learning}, volume~37 of \emph{Proceedings of Machine Learning Research}, pp.\
   448--456. PMLR, 07--09 Jul 2015.

\bibitem[Izmailov et~al.(2018)Izmailov, Podoprikhin, Garipov, Vetrov, and
  Wilson]{AverageWeights}
Izmailov, P., Podoprikhin, D., Garipov, T., Vetrov, D.~P., and Wilson, A.~G.
\newblock {Averaging Weights Leads to Wider Optima and Better Generalization}.
\newblock In \emph{Conference on Uncertainty in Artificial Intelligence}, 2018.

\bibitem[Keskar et~al.(2017)Keskar, Mudigere, Nocedal, Smelyanskiy, and
  Tang]{LargeBatchTraining}
Keskar, N.~S., Mudigere, D., Nocedal, J., Smelyanskiy, M., and Tang, P. T.~P.
\newblock {On Large-Batch Training for Deep Learning: Generalization Gap and
  Sharp Minima}.
\newblock In \emph{International Conference on Learning Representations}, 2017.

\bibitem[Kingma et~al.(2015)Kingma, Salimans, and Welling]{VariationalDropout}
Kingma, D.~P., Salimans, T., and Welling, M.
\newblock {Variational Dropout and the Local Reparameterization Trick}.
\newblock In \emph{Advances in Neural Information Processing Systems 28}, pp.\
  2575--2583. Curran Associates, Inc., 2015.

\bibitem[Krizhevsky(2009)]{CIFAR}
Krizhevsky, A.
\newblock {Learning Multiple Layers of Features from Tiny Images}.
\newblock 2009.

\bibitem[Langford \& Caruana(2002)Langford and Caruana]{NotBounding}
Langford, J. and Caruana, R.
\newblock {(Not) Bounding the True Error}.
\newblock In \emph{Advances in Neural Information Processing Systems 14}, pp.\
  809--816. MIT Press, 2002.

\bibitem[Lecun et~al.(1998)Lecun, Bottou, Bengio, and Haffner]{LeNet}
Lecun, Y., Bottou, L., Bengio, Y., and Haffner, P.
\newblock {{Gradient-based Learning Applied to Document Recognition}}.
\newblock In \emph{Proceedings of the IEEE}, pp.\  2278--2324, 1998.

\bibitem[LeCun et~al.(1998)LeCun, Cortes, and Burges]{MNIST}
LeCun, Y., Cortes, C., and Burges, C. J.~C.
\newblock {The MNIST Database of Handwritten Digits}.
\newblock 1998.

\bibitem[Li et~al.(2018)Li, Xu, Taylor, and Goldstein]{VisualizeLoss}
Li, H., Xu, Z., Taylor, G., and Goldstein, T.
\newblock {Visualizing the Loss Landscape of Neural Nets}.
\newblock In \emph{Advances in Neural Information Processing Systems 31}, pp.\
  6391--6401. Curran Associates, Inc., 2018.

\bibitem[Liang et~al.(2017)Liang, Poggio, Rakhlin, and Stokes]{FisherRaoNorm}
Liang, T., Poggio, T.~A., Rakhlin, A., and Stokes, J.
\newblock {Fisher-Rao Metric, Geometry, and Complexity of Neural Networks}.
\newblock \emph{CoRR}, abs/1711.01530, 2017.

\bibitem[Loshchilov \& Hutter(2019)Loshchilov and Hutter]{AdamW}
Loshchilov, I. and Hutter, F.
\newblock {Decoupled Weight Decay Regularization}.
\newblock In \emph{International Conference on Learning Representations}, 2019.

\bibitem[McAllester(1999)]{SomePACTheorems}
McAllester, D.~A.
\newblock {Some PAC-Bayesian Theorems}.
\newblock \emph{Machine Learning}, 37\penalty0 (3):\penalty0 355--363, Dec
  1999.
\newblock ISSN 1573-0565.

\bibitem[McAllester(2003)]{PACBayesStochasticModelSel}
McAllester, D.~A.
\newblock {PAC-Bayesian Stochastic Model Selection}.
\newblock \emph{Machine Learning}, 51\penalty0 (1):\penalty0 5--21, Apr 2003.
\newblock ISSN 1573-0565.

\bibitem[Neyshabur et~al.(2017)Neyshabur, Bhojanapalli, Mcallester, and
  Srebro]{ExploreGeneralization}
Neyshabur, B., Bhojanapalli, S., Mcallester, D., and Srebro, N.
\newblock {Exploring Generalization in Deep Learning}.
\newblock In \emph{Advances in Neural Information Processing Systems 30}, pp.\
  5947--5956. Curran Associates, Inc., 2017.

\bibitem[Neyshabur et~al.(2018)Neyshabur, Bhojanapalli, and
  Srebro]{PACBayesianSpectralMargin}
Neyshabur, B., Bhojanapalli, S., and Srebro, N.
\newblock {A {PAC}-Bayesian Approach to Spectrally-Normalized Margin Bounds for
  Neural Networks}.
\newblock In \emph{International Conference on Learning Representations}, 2018.

\bibitem[Neyshabur et~al.(2019)Neyshabur, Li, Bhojanapalli, LeCun, and
  Srebro]{OverParam}
Neyshabur, B., Li, Z., Bhojanapalli, S., LeCun, Y., and Srebro, N.
\newblock {The role of over-parametrization in generalization of neural
  networks}.
\newblock In \emph{International Conference on Learning Representations}, 2019.

\bibitem[Rissanen(1986)]{MDL}
Rissanen, J.
\newblock Stochastic complexity and modeling.
\newblock \emph{Ann. Statist.}, 14\penalty0 (3):\penalty0 1080--1100, 09 1986.

\bibitem[Salimans \& Kingma(2016)Salimans and Kingma]{WeightNorm}
Salimans, T. and Kingma, D.~P.
\newblock {Weight Normalization: A Simple Reparameterization to Accelerate
  Training of Deep Neural Networks}.
\newblock In \emph{Advances in Neural Information Processing Systems 29}, pp.\
  901--909. Curran Associates, Inc., 2016.

\bibitem[Sun et~al.(2019)Sun, Zhang, Shi, and Grosse]{FunctionalBBN}
Sun, S., Zhang, G., Shi, J., and Grosse, R.
\newblock {Functional Variational Bayesian Neural Networks}.
\newblock In \emph{International Conference on Learning Representations}, 2019.

\bibitem[Szegedy et~al.(2014)Szegedy, Zaremba, Sutskever, Bruna, Erhan,
  Goodfellow, and Fergus]{Intriguing}
Szegedy, C., Zaremba, W., Sutskever, I., Bruna, J., Erhan, D., Goodfellow, I.,
  and Fergus, R.
\newblock {Intriguing properties of neural networks}.
\newblock In \emph{International Conference on Learning Representations}, 2014.

\bibitem[Tishby et~al.(1999)Tishby, Pereira, and Bialek]{InformationBottleneck}
Tishby, N., Pereira, F.~C., and Bialek, W.
\newblock The information bottleneck method.
\newblock In \emph{Proceedings of the 37-th Annual Allerton Conference on
  Communication, Control and Computing}, pp.\  368--377, 1999.

\bibitem[Wang et~al.(2018)Wang, Shirish~Keskar, Xiong, and Socher]{IdentifyGen}
Wang, H., Shirish~Keskar, N., Xiong, C., and Socher, R.
\newblock {Identifying Generalization Properties in Neural Networks}.
\newblock \emph{ArXiv e-prints}, 2018.

\bibitem[Xie et~al.(2017)Xie, Girshick, Doll{\'{a}}r, Tu, and He]{ResNext}
Xie, S., Girshick, R.~B., Doll{\'{a}}r, P., Tu, Z., and He, K.
\newblock {Aggregated Residual Transformations for Deep Neural Networks}.
\newblock In \emph{2017 {IEEE} Conference on Computer Vision and Pattern
  Recognition}, pp.\  5987--5995, 2017.

\bibitem[Yao et~al.(2018)Yao, Gholami, Lei, Keutzer, and
  Mahoney]{HessianAnalysis}
Yao, Z., Gholami, A., Lei, Q., Keutzer, K., and Mahoney, M.~W.
\newblock {Hessian-based Analysis of Large Batch Training and Robustness to
  Adversaries}.
\newblock In \emph{Advances in Neural Information Processing Systems 31}, pp.\
  4954--4964. Curran Associates, Inc., 2018.

\bibitem[Zagoruyko \& Komodakis(2016)Zagoruyko and Komodakis]{WideResNet}
Zagoruyko, S. and Komodakis, N.
\newblock {Wide Residual Networks}.
\newblock In \emph{Proceedings of the British Machine Vision Conference}, pp.\
  87.1--87.12, 2016.

\bibitem[Zhang et~al.(2019)Zhang, Wang, Xu, and
  Grosse]{ThreeMechanismWeightDecay}
Zhang, G., Wang, C., Xu, B., and Grosse, R.
\newblock {Three Mechanisms of Weight Decay Regularization}.
\newblock In \emph{International Conference on Learning Representations}, 2019.

\bibitem[Zhou et~al.(2019)Zhou, Veitch, Austern, Adams, and
  Orbanz]{NonvacuousImageNet}
Zhou, W., Veitch, V., Austern, M., Adams, R.~P., and Orbanz, P.
\newblock {Non-vacuous Generalization Bounds at the ImageNet Scale: a
  {PAC}-Bayesian Compression Approach}.
\newblock In \emph{International Conference on Learning Representations}, 2019.

\bibitem[Zoph et~al.(2018)Zoph, Vasudevan, Shlens, and Le]{NASNet}
Zoph, B., Vasudevan, V., Shlens, J., and Le, Q.~V.
\newblock {Learning Transferable Architectures for Scalable Image Recognition}.
\newblock In \emph{2018 {IEEE} Conference on Computer Vision and Pattern
  Recognition}, pp.\  8697--8710, 2018.

\end{thebibliography}
  \bibliographystyle{icml/icml2019}

\ificml
\else

  \clearpage

  \appendix

  \section{Running examples}
  \label{sec:appendix:running-examples}

    \subsection{Row and column scaling}
    \label{sec:appendix:running-examples:row-column-scaling}

      We show running examples of the transformation proposed in \Sec{sec:normalized-flat-minima:insufficiency}.
      We consider the following network.
      \begin{align}
        f(X) &= W^{(2)}(\mathrm{ReLU}(W^{(1)}(X))),\\
        W^{(1)} &= \left({\begin{matrix}
               1 & 2\\
               3 & 4\\\end{matrix}}\right),\\
        W^{(2)} &= \left({\begin{matrix}
               5 & 6\\
               7 & 8\\\end{matrix}}\right).
      \end{align}
      Now, matrix norms are as follows.
      \begin{align}
        \frobenius{W^{(1)}} &= \sqrt{30},\\
        \spectral{W^{(1)}} &\approx 5.48,\\
        \frobenius{W^{(2)}} &= \sqrt{174},\\
        \spectral{W^{(2)}} &\approx 13.19.
      \end{align}
      We apply the transformation to the first row of $W^{(1)}$ and the first column of $W^{(2)}$ with $\alpha = 10$.
      Then, parameters change as follows.
      \begin{align}
        W^{(1)} &= \left({\begin{matrix}
               0.1 & 0.2\\
               3 & 4\\\end{matrix}}\right),\\
        W^{(2)} &= \left({\begin{matrix}
               50 & 6\\
               70 & 8\\\end{matrix}}\right).
      \end{align}
      Next, we apply the transformation to the second row of $W^{(1)}$ and the second column of $W^{(2)}$ with $\alpha = 0.1$.
      Parameters change as follows.
      \begin{align}
        W^{(1)} &= \left({\begin{matrix}
               0.1 & 0.2\\
               30 & 40\\\end{matrix}}\right),\\
        W^{(2)} &= \left({\begin{matrix}
               50 & 0.6\\
               70 & 0.8\\\end{matrix}}\right).
      \end{align}
      Now, matrix norms changed as follows.
      \begin{align}
        \frobenius{W^{(1)}} &= \sqrt{2500.05},\\
        \spectral{W^{(1)}} &\approx 50.00,\\
        \frobenius{W^{(2)}} &= \sqrt{6101.13},\\
        \spectral{W^{(2)}} &\approx 78.11.
      \end{align}
      Using the same method, we can make matrix norms of both $W^{(1)}$ and $W^{(2)}$ arbitrarily large.

  \section{Convexity of variance parameters}
  \label{sec:appendix:convexity}

    We show that the optimization problem\,\eqref{eq:normalized-flat-minima} is convex with respect to the log of variance parameters $\vec{\sigma}^2$ and $\vec{\sigma}'^2$.
    Let us define the following parameters.
    \begin{align}
      e^{\alpha_i} &:= \sigma_i^2, \\
      e^{\beta_j} &:= \sigma_j'^2.
    \end{align}
    For the sake of notational simplicity, we rewrite the objective in \eqref{eq:normalized-flat-minima} as follows.
    \begin{equation}
      S := \sum_{i,j}\left( a_{i,j}e^{\alpha_i}e^{\beta_j} + b_{i,j}e^{-\alpha_i}e^{-\beta_j}\right),
    \end{equation}
    where $a_{i,j} \geq 0$ and $b_{i,j} \geq 0$ for all $i$ and $j$.
    To show that $S$ is convex with respect to $\vec{\alpha}$ and $\vec{\beta}$, we show that the Hessian of $S$ is semi-positive definite.
    First, we calculate the elements of the Hessian.
    \begin{align}
      \frac{\partial^2S}{\partial \alpha_i\partial \alpha_j} &= \sum_{k}\left( a_{i,k}e^{\alpha_i}e^{\beta_k} + b_{i,k}e^{-\alpha_i}e^{-\beta_k}\right)\delta_{i,j},\\
      \frac{\partial^2S}{\partial \beta_i\partial \beta_j} &= \sum_{k}\left( a_{k,j}e^{\alpha_k}e^{\beta_j} + b_{k,j}e^{-\alpha_k}e^{-\beta_j}\right)\delta_{i,j},\\
      \frac{\partial^2S}{\partial \alpha_i\partial \beta_j} &= a_{i,j}e^{\alpha_i}e^{\beta_j} + b_{i,j}e^{-\alpha_i}e^{-\beta_j}.
    \end{align}
    For the notational simplicity, we define the following.
    \begin{equation}
      \gamma_{i,j} := a_{i,j}e^{\alpha_i}e^{\beta_j} + b_{i,j}e^{-\alpha_i}e^{-\beta_j}.
    \end{equation}
    Note that $\gamma \geq 0$.
    We can rewrite the elements of the Hessian as follows.
    \begin{align}
      \frac{\partial^2S}{\partial \alpha_i\partial \alpha_j} &= \sum_{k}\gamma_{i,k}\delta_{i,j},\\
      \frac{\partial^2S}{\partial \beta_i\partial \beta_j} &= \sum_{k}\gamma_{k,j}\delta_{i,j},\\
      \frac{\partial^2S}{\partial \alpha_i\partial \beta_j} &= \gamma_{i,j}.
    \end{align}
    Now, it sufficies to show that $\forall v, v^\top \left(\nabla_{\vec{x},\vec{\beta}}^2 S\right) v \geq 0$.
    \begin{align}
      &v^\top \left(\nabla_{\vec{\sigma},\vec{\sigma}'}^2 S\right) v\\
      =& \sum_{i,j}(\gamma_{i,j}v_{\alpha_i}^2) + \sum_{i,j}(\gamma_{i,j}v_{\beta_j}^2) + 2\sum_{i,j}\gamma_{i,j}v_{\alpha_i}v_{\beta_j}\\
      =& \sum_{i,j}\gamma_{i,j}(v_{\alpha_i} + v_{\beta_j})^2 \geq 0 \qed
    \end{align}

  \section{Calculation of normalized sharpness}
  \label{sec:appendix:gd}

    To calculate the normalized flat minima\,\eqref{eq:normalized-flat-minima}, we have to solve the following optimization problem for each weight matrix.
    \begin{align}
      \min_{\vec{\sigma},\vec{\sigma'}} \sum_{i,j}\left( \frac{\partial^2\loss{\train}{\hyp_\param}}{\partial W_{i,j}\partial W_{i,j}}(\sigma_i\sigma'_j)^2 + \frac{W_{i,j}^2}{2\lambda(\sigma_i\sigma'_j)^2}\right).
    \end{align}
    The parameter $\lambda$ is arbitrary but we set $\lambda = 1$ for simplicity.
    If we can estimate the diagonal elements of the Hessian, the later parts are straightforward.
    We can use the following to estimate the diagonal elements of the Hessian.
    \begin{align}
      &\expect{\vec{\epsilon}\sim\mathcal{N}(\vec{0},\vec{1})}{
        \vec{\epsilon}\odot\nabla_\param^2\loss{\train}{\hyp_\param}\vec{\epsilon}
      }\\
      \approx &
      \expect{\vec{\epsilon}\sim\mathcal{N}(\vec{0},\vec{1})}{
        \vec{\epsilon}\odot
        \frac{\nabla_{\param} \loss{\train}{\hyp_{\param + r\vec{\epsilon}}} -
        \nabla_{\param} \loss{\train}{\hyp_{\param - r\vec{\epsilon}}}}{2r}
      },
    \end{align}
    where $r>0$ is a small constant.
    In our experiments, $r$ was chosen per weight matrix according to their Frobenius norm for better estimation of the Hessian.

  \section{Alternative definition of the normalized sharpness}
  \label{sec:appendix:alternative}

    As an alternative to \eqref{eq:normalized-flat-minima}, we can directly solve the following optimization problem in \eqref{eq:pac-bayes-D}.
    \begin{equation}
      \label{eq:normalized-flat-minima-by-noise}
      \min_{\vec{\sigma},\vec{\sigma'}}\left(\left(\loss{\train}{\posterior_{\vec{\sigma},\vec{\sigma}'}} - \loss{\train}{\hyp}\right) + \sum_{i,j,k}\frac{(W^{(k)}[i,j])^2}{2\lambda(\sigma^{(k)}_i\sigma_j'^{(k)})^2}\right).
    \end{equation}
    We can use stochastic gradient descent to optimize $\vec{\sigma}$ and $\vec{\sigma}'$.
    An advantage of \eqref{eq:normalized-flat-minima-by-noise} over \eqref{eq:normalized-flat-minima} is that we do not need a second-order approximation.
    However, we have the following disadvantages.
    \begin{itemize}
      \item The optimization problem becomes nonconvex.
      \item Sharpness becomes sensitive to the choice of $\lambda$.\footnote{The
        quantity\,\eqref{eq:normalized-flat-minima} is invariant to the choice of $\lambda$ in a sense that sharper model is always sharper no matter which $\lambda$ we set. On the other hand, the quantity\,\eqref{eq:normalized-flat-minima-by-noise} does not have the invariance.
      }
    \end{itemize}
    Given these disadvantages of \eqref{eq:normalized-flat-minima-by-noise}, we use \eqref{eq:normalized-flat-minima} as a definition of the normalized sharpness.

  \section{Experimental setups}
  \label{sec:appendix:setup}

    \subsection{Setups of \Sec{sec:experiments:nf-rl}}
    \label{sec:appendix:setup:1}

      Ratio of random labels were selected from ($0.0$, $0.1$, $0.2$, $0.3$, $0.4$, $0.5$, $0.6$, $0.7$, $0.8$, $0.9$, $1.0$) uniform randomly at each training.
      We used cross-entropy loss during the training.
      We used a normalized-softmax-cross-entropy loss\,\eqref{eq:nsce} to calculate normalized sharpness.

      \paragraph{MLP on MNIST:}
      We trained MLP for 50 epochs with batchsize 128 on MNIST.
      We used Adam optimizer with its default parameters ($\mathrm{lr} = 0.001$).

      \paragraph{LeNet on MNIST:}
      We trained LeNet for 50 epochs with batchsize 128 on MNIST.
      We used Adam optimizer with its default parameters ($\mathrm{lr} = 0.001$).

      \paragraph{LeNet on CIFAR10:}
      We trained LeNet for 100 epochs with batchsize 128 on CIFAR10.
      We used Adam optimizer with its default parameters ($\mathrm{lr} = 0.001$).

      \paragraph{Wide ResNet on CIFAR10:}
      We trained $16$ layers Wide ResNet for 200 epochs with batchsize 128 on CIFAR10.
      We used width factor $k=4$.
      We used Adam optimizer with its default parameters ($\mathrm{lr} = 0.001$).

    \subsection{Setups of \Sec{sec:experiments:f-rl}}
    \label{sec:appendix:setup:2}

      We used the same setups described in \Sec{sec:appendix:setup:2}.
      We used cross-entropy loss for both training and calculation of the Hessian.

\fi

\end{document}